\documentclass{egpubl}
\usepackage{eurovis2026}

\SpecialIssuePaper         

\CGFccby

\usepackage[T1]{fontenc}
\usepackage{dfadobe}
\usepackage{todonotes}

\usepackage{cite}  
\BibtexOrBiblatex
\electronicVersion
\PrintedOrElectronic
\ifpdf 
\else \usepackage[dvips]{graphicx} \fi

\usepackage{egweblnk}

\usepackage{tabularx}
\usepackage{graphicx} 
\usepackage{siunitx}
\usepackage{booktabs}
\usepackage[utf8]{inputenc}
\usepackage{array} 
\usepackage{booktabs} 
\usepackage[table]{xcolor}
\usepackage{xspace}
\usepackage{makecell}
\usepackage{lineno}
\usepackage{todonotes}
\usepackage{amsmath} 
\usepackage{amssymb}
\usepackage{multirow}

\usepackage{xcolor, colortbl, booktabs}

\usepackage{tablefootnote}
\usepackage{tcolorbox}

\graphicspath{{figures/}{pictures/}{images/}{./}} 

\usepackage{times}                     

\usepackage{tabu}                      
\usepackage{booktabs}                  
\usepackage{lipsum}                    
\usepackage{mwe}                       

\usepackage{mathptmx}                  
\usepackage{xcolor}
\usepackage{marginnote}

\usepackage{xr}          
\usepackage[normalem]{ulem}

\newcommand{\yifan}[1]{{\color{red} #1}}

\newcommand{\revision}[1]{{\color{black}#1}}
\newcommand{\removed}[1]{}

\title[Beauty in the Eye of AI]{Beauty in the Eye of AI: Aligning LLMs and Vision Models with Human Aesthetics in Network Visualization}

\author[P.Zhang \& X.Li \& X.Wang \& H.Shen \& Y.Hu]
{\parbox{\textwidth}{\centering P.Zhang$^{1}$
X.Li$^{1}$
X.Wang$^{2}$
H.Shen$^{3}$
and Y.Hu$^{1}$
}
\\
{\parbox{\textwidth}{\centering 
$^1$ Northeastern University, United States\\
$^2$ Bosch AI Research, United States \\
$^3$ The Ohio State University, United States
  }
}
}

\begin{document}

\maketitle

\begin{abstract}
Network visualization has traditionally relied on heuristic metrics, such as stress, under the assumption that optimizing them leads to aesthetic and informative layouts. However, no single metric consistently produces the most effective results. A data-driven alternative is to learn from human preferences, where \revision{lablers}\removed{annotators} select their favored visualization among multiple layouts of the same graphs. These human-preference labels can then be used to train a generative model that approximates human aesthetic preferences. However, obtaining human labels at scale is costly and time-consuming. As a result, this generative approach has so far been tested only with machine-labeled data~\cite{wang_2023_smartgd}. In this paper, we explore the use of large language models (LLMs) and vision models (VMs) as proxies for human judgment. Through a carefully designed user study involving 27 participants, we curated a large set of human preference labels. We used this data both to better understand human preferences and to bootstrap LLM/VM labelers. We show that prompt engineering that combines few-shot examples and diverse input formats, such as image embeddings, significantly improves LLM–human alignment, and additional filtering by the confidence score of the LLM pushes the alignment to human–human levels. Furthermore, we demonstrate that carefully trained VMs can achieve VM-human alignment at a level comparable to that between human \revision{lablers}\removed{annotators}. Our results suggest that AI can feasibly serve as a scalable proxy for human labelers.
\end{abstract}

\section{Introduction}

\begin{figure}[htbp!]
\vspace{-0.3cm}
        \includegraphics[width=0.5\textwidth]{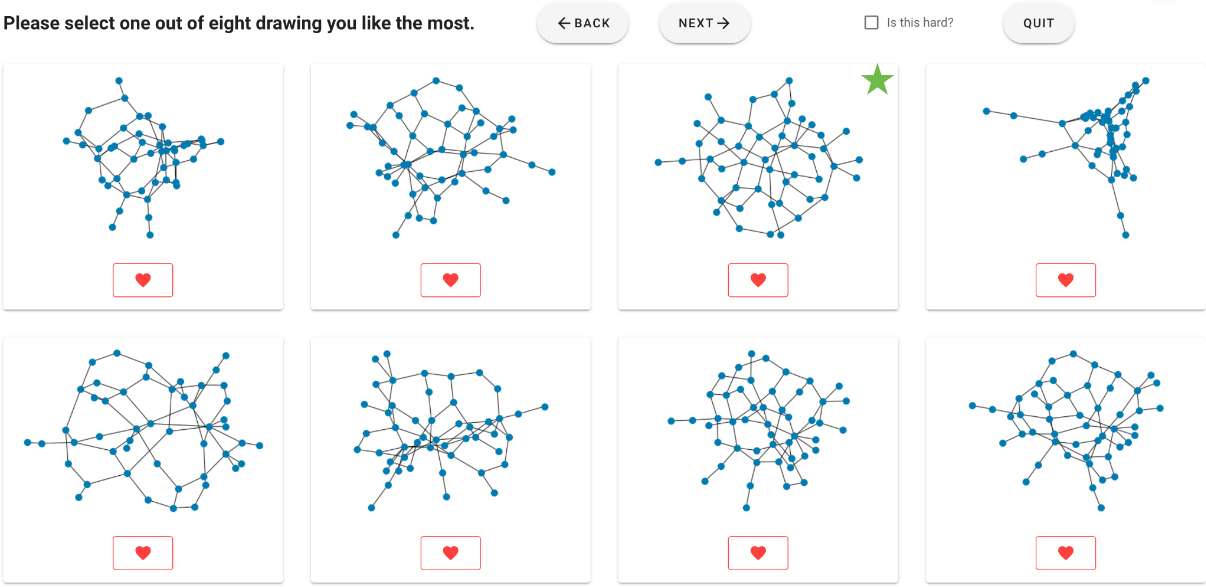} 
\vspace{-0.2cm}
    \caption{A UI showing eight graph visualizations. Of the users ($n=43$), 56\% preferred the starred option, and 16\% chose the bottom-left. }
    \label{fig:data_collection}
\vspace{-0.4cm}
\end{figure}

Graphs are ubiquitous in many applications, including social, biological,  transport, and computer networks. 
Graph visualization transforms complex networks into intuitive graphics that make information accessible and comprehensible. 
Current network visualization algorithms often optimize researcher-defined aesthetic metrics that are presumed to reflect user preferences, yet real user preferences may only partially align with these metrics. As an example, Fig.~\ref{fig:data_collection}(a) shows different visualizations of the {\em same} graph, created by eight popular network visualization algorithms, each optimizing an aesthetic criterion. When we asked a group ($n=43$) of people, many of them favored the starred visualization, yet some preferred other options. This suggests that humans tend to prefer certain kinds of visualization, but their preferences are also diverse. Thus far, we don't fully understand what humans are looking for, let alone have an analytical formula to describe it.

Rather than relying on presumptive aesthetic metrics, we argue for a more human-centric approach to network visualization. 
We envision that this can be achieved by training a deep learning model to learn human preferences directly. This requires
collecting human-labeled examples of ``good'' and ``bad'' visualizations and training a discriminator that closely reflects human aesthetic judgments and task performance. Such a discriminator could then guide the automatic generation of visualizations that align with human perceptions of beauty and effectiveness in visual-analytic tasks. 
The works of DeepGD and SmartGD~\cite{deepgd, wang_2023_smartgd} have shown that with a GAN framework, it is possible to infer the aesthetic preference of ``human-proxy'' labelers and subsequently generate visualizations that optimize the learned aesthetic preferences. In these works, since collecting human labels is expensive, the authors used a ``mechanical labeler'' that chose among multiple visualizations one that minimizes a known metric (e.g., edge crossing). 

In this paper, we take the next step toward human-centric generative visualization by collecting a large number of actual human labels. To facilitate this, we developed a custom user interface (Figure~\ref{fig:data_collection}) and conducted a carefully designed human study involving 27 participants. In this study, participants were asked to select their preferred visualization from eight layouts generated by eight widely used graph layout algorithms. In total, we collected human-preference labels for 11,531 graphs, amounting to 64,436 labels (each label corresponds to a unique graph–annotator pair).
This process underscored the significant cost of large-scale human annotation -- it required 172 person-hours over many months. {\em Such a high cost of human labeling limits the ability to further scale up data collection}, including incorporating more graphs, additional algorithms, and perturbed variants of existing layouts.

With the recent advent of Large Language Models (LLMs), particularly Vision-Language Models (VLMs), a new possibility has emerged: using an LLM to emulate human preferences. In parallel, advances in vision models (VMs), especially self-supervised vision foundation models (VFMs) such as DINOv2 that do not rely on handcrafted labels, have greatly improved the ability to analyze and understand images. In this paper, we address the following research question with regard to network visualization:

\begin{itemize}
    \item \textbf{RQ1}: Do human \revision{lablers}\removed{annotators} share similar aesthetic preferences?  
    \item \textbf{RQ2}: Do Large Language Models (LLMs) understand human aesthetic preferences, or can they be guided to align with human?  
    \item \textbf{RQ3}: Can Vision Models (VMs) learn from human labels and align with human aesthetic preferences?  
\end{itemize}

Regarding \textbf{RQ1}, our user study revealed that humans share common aesthetic preferences, yet also exhibit substantial diversity. Due to this diversity, and the fact that several top graph layout algorithms often produce visually similar layouts, any two human \revision{lablers}\removed{annotators} agreed with each other in 38.34\% of cases on average.

Regarding \textbf{RQ2}, we found that prompt engineering is crucial for enhancing the image-understanding capabilities of LLMs. Simply asking an LLM to select the best visualization from a set of layout images was ineffective. To address this, we introduced a process that leverages an image-embedding–based memory bank, which improved the alignment between LLM-generated preferences and human \revision{lablers}\removed{annotators}’ choices by up to ~13\% compared with the naïve image-only approach. These improvements bring LLM–human alignment close to the level of human–human agreement.  With token-probability filtering, LLM–human alignment can match human–human alignment.

Regarding \textbf{RQ3}, we found that with a carefully designed training strategy that accounts for the diversity of human preferences, and the use of a state-of-the-art Vision Foundation Model (VFM), we are able to achieve VFM–human alignment comparable to human–human alignment. 

We believe these results open up the possibility of greatly reducing the cost of label collection and may ultimately enable truly human-centric generative graph visualization. The main contributions of this papers are as follows:
\begin{itemize}
    \item We design and conduct a large-scale user study that collects human-preferred visualizations for 11,531 graphs (64,436 labels), creating a \revision{benchmark}\revision{dataset} that directly captures human aesthetic preferences. 
    \item We develop an LLM-based preference modeling pipeline that uses prompt engineering and an embedding-based memory mechanism to improve the model’s ability to replicate human aesthetic judgments. With label confidence control, an LLM can serve as an effective proxy for human labelers.
    \item We propose a VM-based learning framework that incorporates human label diversity and trains a VFM to predict layout preferences, providing a strong model aligned with human judgments.
\end{itemize}

The dataset, together with the code used in our experiments, will be released as open source upon publication of this paper.

\section{Related work}

\subsection{Graph aesthetic metrics and layout algorithms}


A widely used approach for generating straight-line drawings involves physical models that minimize the energy of the system~\cite{Eades_1984, Fruchterman_Reigold_1991, kamada_kawai_1989, neato, zheng-gd2}, optimized through force-directed algorithms\cite{kobourov_2013}. Other layout methods specifically target aesthetic metrics, such as minimizing edge crossings~\cite{xing-heuristic, spx},  maximizing crossing angles~\cite{Argyriou, Bekos-xangle}, shape-metric~\cite{Eades2017ShapeMetrics, sgd2}, or a combination\cite{didimo, sgd2, wang_2023_smartgd}.


To evaluate the quality of network layouts, various aesthetic criteria were proposed in the literature (e.g., stress, edge crossings, crossing angles, node distribution, and node occlusion), often based on intuition, or on insights from Gestalt principles. 
User studies~\cite{purchase1, purchase_2011_gdaesthetics} have confirmed that some of these criteria do correlate with human preferences of graph layouts~\cite{Tim-user-study}. However, each criterion addresses only one aspect of aesthetics, and they may conflict with one another~\cite{haleem-huamin,spx}. For example, algorithms that optimize crossing angles often result in layouts with high stress and poor neighborhood preservation~\cite{spx}. 
Experiments~\cite{Chimani_2014_stress,Mooney_2024_stress} showed that when forced to choose between two graphical representations,  participants preferred the drawing with lower stress (57\%) or fewer crossings (65\%). Hence, neither is preferred 100\% and can be a proxy of human preference.
There is also no consensus on how to best combine multiple criteria to model human preferences. This motivated us to conduct a user study to curate a large set human labels, with the goal of using these labels to understand human preferences and to bootstrap the process of collecting even more labels through human-preference-aligned LLMs or VMs.


\subsection{Using deep learning to evaluate graph layout metrics or human preference}

Tiezzi et al.\cite{gnn-gd}  proposed to use a simple MLP with two hidden layers\ to predict whether two edges cross. Haleem et al.~\cite{haleem-huamin} involves training CNN models to predict metrics such as node occlusion, edge crossings, and minimum crossing angles from images. Klammler et al.~\cite{Klammler2018GD} considered modeling aesthetics using a two-layer neural network. Their labeled dataset consists of 36K layout pairs. Good layouts were constructed using FM${}^3$~\cite{Hachul_Junger_2004} and stress minimization. Bad ones
where created by perturbation or random layouts. No human labels were involved.



Cai et al.~\cite{Cai2021PacificVis} proposed a CNN-based machine learning approach for predicting human preferences for graph layouts. They used a small set of 146 graphs, and the task was a pairwise comparison of layouts with 511 human-labeled pairs; therefore, they appropriately used a Siamese CNN model. Because of the small dataset, they first trained the network using a majority vote of three metrics (shape, crossings, stress) as labels and then applied transfer learning by fine-tuning with human-labeled data. A similar study, but using graph neural networks (GNNs) instead of CNNs, was conducted by Wu et al.~\cite{Wu2025GNNGraph}, where the goals were either to predict existing metrics (e.g., edge crossings) or to predict which of two layouts a human would prefer. The study used 800 pairs of manually annotated pairs. Our work differs from these in that our task is to select one layout out of eight, and our overall dataset is two orders of magnitude larger. For our VM-based approach, we treat this as a multiclass classification problem, embedding each of the eight images using a vision models (ResNet) and a vision foundation model (DINOv2/3). 
We do not pretrain by assuming that human preferences align with any particular metric, as in~\cite{Cai2021PacificVis}. Our prediction task (one out of eight) is more challenging, the human-labeled dataset we curated is orders of magnitude larger, and we investigate both VM- and LLM-based approaches.

\subsection{Deep learning approaches for graph drawing}

In recent years, there has been increased interest in developing deep learning-based graph drawing methods~\cite{kwon-ma-2020,dnn,gnn-gd,kwon-ma-2020,deep-drawing,deepgd,wang2020deepdrawing}. 
While earlier works often lack generalizability, $(DNN)^2$~\cite{dnn}, which employs a Graph Convolution Network (GCN), is generalizable but limits input graphs to a maximum size of 120.
In contrast, DeepGD~\cite{deepgd} can handle any differentiable loss function and, once trained, can be applied to graphs of any type. However all these algorithms require a differential objective function.

For non-differentiable objective function, Ahmed et al.\cite{sgd2} proposed a flexible framework called $\mathrm{GD}^2$, which uses stochastic gradient descent to optimize layouts for any differentiable criterion. This method can only handle non-differentiable criteria through hand-crafted surrogate functions. 

SmartGD~\cite{wang_2023_smartgd} moved away from an explicit definition of a cost function, and requires only examples of ``good'' drawings for a set of training graphs rather than a quantifiable measure of goodness. 
It demonstrated that a GAN-based approach has the potential to learn from examples only. The logical next step is to guide such a generative visualization algorithm through a large amount of human-labelled ``good'' visualizations. 
However, collecting a large number of human-preferred visualizations is expensive, motivating us to explore the use of LLMs and VMs as proxies for human \revision{lablers}\removed{annotators}, once they are aligned with human preferences.

\subsection{LLM and visualization}

LLMs offer new opportunities for visualizations, but at the same time raise new questions~\cite{ScDiEl+2023Doom-1}. A multitude of works are emerging in applying generative AI in visualization~\cite{ye2024generative}. One natural application is in NL2Vis, for example ADVISor~\cite{liu2021advisor} trained neural networks for NL2SQL as well as a rule-based visualization generation step. Data2Vis~\cite{dibia2019data2vis} converts natural language descriptions of data table columns into Vega-Lite code sequences using a sequence-to-sequence RNN. 

Our work is concerned with network visualization, in particular, with aligning the aesthetic preference of LLMs with that of human beings. In that area, Wang et al.~\cite{wang2025aligned} considered whether LLMs can predict human takeways of 4 different types of bar charts, and found LLMs exhibit variations in takeaways of the same chart type that were not seen in human subjects. In the graph visualization area, Bartolomeo et al.~\cite{DiBartolomeo2023AskYouShall} investigated the application of ChatGPT in implementing steps of layered graph drawing algorithms and found that the LLM produced encouraging results on some sub-tasks, but is overall limited in capability.

\section{Human Preference Dataset}
\label{sec:dataset}

To quantitatively evaluate the extent to which LLMs or VMs can serve as proxies for human judgment, it is essential to establish a dataset grounded in human preferences. To this end, we conducted an extensive within-subject user study. Participants were asked to select the most aesthetically pleasing visualization from eight \removed{canonical}\revision{most popular} layout algorithms applied to the 11,531 graphs from the entire Rome Graphs Collection~\cite{rome_graphs}. The eight algorithms are \texttt{Neato}~\cite{neato},  \texttt{Kamada-Kawai (KK)}~\cite{kamada_kawai_1989}, \texttt{FA2}~\cite{fa2}, \texttt{fdp}~\cite{graphviz}, \texttt{sfdp}~\cite{sfdp},
\texttt{spring}~\cite{spring}, \texttt{PMDS}~\cite{pmds}, and \texttt{spectral}~\cite{spectral}. To avoid introducing bias toward any predefined aesthetic criteria (e.g., edge crossings), participants were instructed only to ``select one drawing you like the most''. This enables us to capture the abstract human preference without injecting bias from pre-defined graph drawing aesthetic criteria.

The study followed three principles: \emph{scalability} (to annotate thousands of graphs), \emph{coverage and redundancy} (via repeated labeling), and \emph{conflict resolution} (to address subjective disagreement). Participants interacted through a web interface (Figure~\ref{fig:data_collection}) that displayed all eight layouts simultaneously in a randomized, unlabeled grid to eliminate position bias. For each graph, users clicked a heart button to select their preferred layout, optionally marked the decision as difficult, and advanced to the next task. Decision time was recorded as a proxy for cognitive effort. The interface was designed for accessibility and low cognitive load, enabling participation from both novices and experts.

New users completed a brief tutorial explaining the task and UI controls.
Progress feedback and motivational messages (e.g., ``You have labeled more graphs than 85.71\% of users!'') encouraged sustained engagement. An adaptive assignment system prioritized graphs with zero or one label, then those with conflicting labels, ensuring broad coverage and consensus. Skipped graphs were stored in a personal queue and resurfaced with 40\% probability


We recruited \removed{27}\revision{25} participants (\removed{25}\revision{23} novices, 2 with visualization background, and 2 experts. \revision{We excluded results from two labelers whose agreement with the average labelers fell below 0.25}), all with at least a BSc degree, and conducted the study from May 2022 to August 2025. \revision{We removed results from two lablers due to their alignment with average lablers below 25\%}. They provided a total of \revision{64{,}222}\removed{64{,}436} human-preference labels across 11{,}531 graphs (mean 5.58 labels per graph, average \revision{9.59}\removed{9.61}\,s per label) \revision{after data cleaning}. Layouts produced by Kamada--Kawai (\revision{43.4 $\pm$ {0.4} }\removed{44.3}\%) and Neato (34.8 $\pm$ {0.4}\%) were selected most often, whereas spectral layouts were rarely preferred (\revision{0.16 $\pm$ {0.03} }\removed{0.17}\%). Each graph received between 3 and 7 labels.

\subsection{Preference alignment metric}

To formally characterize the human labelers, we first define some notations. Denote $G=\{V, E\}$ as a graph with $V$ the set of nodes and $E$ the set of edges. Let $D$ denote the set of all graphs used for human labeling, and let $D(i)$ be the subset of graphs labeled by labeler $i$. Suppose the visualization chosen for a graph $G$ by labeler $i$ is denoted as $l(G,i)$, where $l(G,i) \in \{1,2,\ldots,8\}$.
We define the pairwise alignment between two labelers $i$ and $j$ as:

\vspace{-0.3cm}
\begin{equation*}
\text{Alignment}(i,j) =
\frac{\sum_{G \in D(i)\cap D(j)} \delta(l(G,i) = l(G,j))}{|D(i)\cap D(j)|},
\end{equation*}
\vspace{-0.3cm}

\noindent where $\delta(\cdot)$ is the indicator function that equals 1 if the condition holds and 0 otherwise.
This measure represents the fraction of graphs on which two labelers make the same choice, effectively the accuracy of one labeler when treating the other’s selections as ground truth.
The overall micro-averaged alignment among all labelers is then defined as:

\vspace{-0.3cm}
\begin{equation}
\text{Alignment} =
\frac{\sum_{i,j} \sum_{G \in D(i)\cap D(j)} \delta(l(G,i) = l(G,j))}{\sum_{i,j} |D(i)\cap D(j)|}.\label{eqn:alignment}
\end{equation}
\vspace{-0.3cm}

We found that human preferences exhibit notable diversity. Pairwise alignment among the seven users with the largest numbers of labels ranged from 25\% to 50\% (Figure~\ref{fig:human_human_llm_vm}), reflecting both the fact that outputs from some algorithms often appear equally good and the inherent variability of individual aesthetic judgments. On average, when two human subjects are randomly selected, their alignment is $38.34\%$, with a macro-average of 38.71\%. While this may appear to be low, it is actually much higher than the alignment of a random choice labeler with an average human labeler, which is calculated as 11.5\%. This indicates that human labelers do share common aesthetic preferences, despite the diversity. 

Our observations are consistent with a study on human preferences for bar-chart aesthetics~\cite{Wu2021CHIChart}, collected via Amazon Mechanical Turk (MTurk), which found that three participants preferred the same chart {\em out of a pair} 45.6\% of the time, compared with 25\% for random choices. Given that our labelers choose 1 out of 8 options instead of 1 out of 2, the human agreement in our study is relatively higher. We believe this supports the quality of our dataset. Consensus analysis (Table~\ref{tab:consensus}) shows that only 5.15\% of graphs achieved unanimous agreement, while 40.79\% resulted in two distinct preferred layouts, highlighting the subjective nature of the task. \revision{We also report the distribution on the test set after replacing a randomly selected human labeler with either an LLM or a VM. For the VM, the distribution remains largely similar to that of all-human labelers, except that the rate of unanimous agreement decreases to 2.02\%. When the LLM is used, the distribution shifts more noticeably.}

\begin{table}[t]
\centering
\caption{Distribution (\%) of unique preferred layouts per graph. \revision{*Results based on the 1{,}000-graph test set.}}
\begin{tabular}{cc|ccc}
\toprule
$\substack{\textbf{Unique} \\ \textbf{Layouts}}$ & \textbf{Human} & \revision{\textbf{Human*}} & \revision{\textbf{incl. LLM*}} & \revision{\textbf{incl. VM*}}\\
\midrule
1 & 5.15  & \revision{4.50}  & \revision{2.95}  & \revision{2.02}  \\
2 & 40.79 & \revision{39.70} & \revision{31.72} & \revision{38.24} \\
3 & 38.55 & \revision{41.5}  & \revision{43.93} & \revision{42.16} \\
4 & 13.49 & \revision{11.8}  & \revision{17.27} & \revision{14.81} \\
5 & 1.94  & \revision{2.50}  & \revision{3.56}  & \revision{2.63}  \\
6 & 0.08  & \revision{0.00}  & \revision{0.56}  & \revision{0.18}  \\
\bottomrule
\end{tabular}
\label{tab:consensus}
\vspace{-0.5cm}
\end{table}

\subsection{Post-study survey}

We also conducted post-study surveys with the participants. For the visualizations of 10 randomly chosen graphs, we asked them to provide free-text explanations for why they preferred a given visualization and why they did not choose the others. We analysed the responses using ChatGPT-4o-mini. We found that the most frequent reasons participants favored a visualization were \emph{symmetry}, \emph{no edge crossings}, \emph{clear structure}, and \emph{uniform spacing/edge lengths}, whereas they disliked \emph{clutter}, \emph{overlaps}, \emph{asymmetry}, and \emph{distortion}. Additional details are provided in the Appendix. \revision{Note that we did not guide the labelers during the user study, and these preferences emerged organically.}

\subsection{Findings for RQ1}
Our user study and subsequent analysis show that humans share common aesthetic preferences: their alignment rate (38.34\%) is substantially higher than random (11.5\%). When visually similar layouts are treated as equivalent, alignment increases further. For example, to 50.67\% when layouts with Procrustes statistics $\ge 0.95$ are considered identical (see Appendix). Nevertheless, the less-than-perfect agreement also highlights the diversity of human preferences. This human-labeled dataset thus provides a strong foundation for learning aesthetic preferences in graph visualization and for bootstrapping automated labeling systems using LLMs or VMs, as discussed next.



\section{Aligning LLM with Human Aesthetic Preferences}

To investigate \textbf{RQ2}, we ask the LLM to select one visualization from a set of eight options for the same graphs and examine whether its choices align with those of human \revision{lablers}\removed{annotators} in the Human Preference Dataset. The key question, then, is {\em how we determine whether the LLM is aligning well with human preferences?} Given that human preferences are diverse, which human labeler should the LLM be compared against?

As we founds in Section~\ref{sec:dataset}, on average, two human labelers agree with each other 38.34\%. If an LLM or VM is to serve competently as a proxy human annotator, it must align with human judgments at a level indistinguishable from that of an average human labeler. This degree of alignment thus defines our target for LLM- or VM-based labeling. 
More precisely, we used the Alignment metric (equation~\ref{eqn:alignment}) to quantify LLM–human or VM–human alignment. In this equation, $i$ denotes the LLM or VM under consideration, and $j$ indexes all human labelers. This metric captures the average alignment of the LLM or VM with the human labelers. {\em We are looking for LLM–human and VM-human alignment close to 38.34\%}.

We experimented with a range of prompts, including 0-shot and few-shot, as well as image-only vs using other features. {\em In all experiments, to avoid LLMs (and VMs later on) selecting a fixed position, the order of the eight layouts is randomly permuted for each test graph.}

\subsection{0-shot vs few-shot image only}

We organized the test data into sets, with each set comprising eight different layouts of a graph. In our 0-shot experiment using only image data with GPT-4o-mini, we carefully designed the prompt following the chain-of-thought prompting technique (Table \ref{tab:prompt_zero_shot_image_only}). 

\begin{table}[htb]
\vspace{-0.3cm}
    \centering
    \caption{Prompt for 0-shot evaluation of layout aesthetics}
    \label{tab:prompt_zero_shot_image_only}
    \begin{tabular}{|p{0.45\textwidth}|}
    \hline
    \small{You are an expert in human aesthetics. I want you to evaluate the graph's layout primarily from an aesthetic perspective. Based on your aesthetic judgment, select the layout that best aligns with human preferences. {\em Let's think step by step}. Give me the answer with the style: "Reason: Result: layout ?" Just give me plain text, do not add emphasize markdown.}\\
    \hline
    \end{tabular}
\vspace{-0.3cm}
\end{table}

We also experimented with a few-shot setup using image-only inputs with GPT-4o-mini. The prompt structure was similar to a zero-shot approach, with the key distinction being the inclusion of shots (eight visualizations as a single set) and their corresponding human-selected optimal layouts (ground truth) within the prompt. The detailed prompt example for the 5-shot configuration is given in Table~\ref{tab:prompt_five_shot_image_only}. 

\begin{table}[h]
\vspace{-0.3cm}
    \centering
    \caption{An example prompt for 5-shot evaluation of layout aesthetics; this is added to the prompt in Table~\ref{tab:prompt_zero_shot_image_only}, just before ``Let's think step by step''.}
    \label{tab:prompt_five_shot_image_only}
    \begin{tabular}{|p{0.45\textwidth}|}
    \hline
    \small{ ... To systematically analyze the layout preferences, we provided multiple exemplar images for layout assessment and human aesthetic preference exploration:

    In the first set of eight layouts, the human selected layout 2.
    In the second set of eight layouts, the human selected layout 7.
    In the third set of eight layouts, the human selected layout 8.
    In the fourth set of eight layouts, the human selected layout 4.
    In the fifth set of eight layouts, the human selected layout 8.
    
    We request a comprehensive analysis of the aesthetic criteria that influenced these layout selections. It is important to note that the preference is independent of the absolute layout numbering. ...}\\
    \hline
    \end{tabular}
\vspace{-0.5cm}
\end{table}

\subsection{Few-shot with graph structural and coordinate features}

We explore alternative approaches that use embeddings to represent graph layouts for prompting LLMs. The prompt format is shown in Table~\ref{tab:embeddings}. Each layout is encoded using two complementary modalities: graph structural information and coordinate features.

The primary goal of this method is to investigate whether geometric and structural information, beyond raw images, can effectively convey layout information to LLMs. When prompted with meaningful graph representations, LLMs may better reason about aesthetic preferences.

\textbf{Graph Structure Information:} This captures the structural topology of the graph through various methods:
\begin{itemize}
    \item \textbf{Edge List:} Edge List represents the graph as a set of node pairs that indicate connections. Formally, the graph is represented as a set of edges $E = \{(u, v) \mid u, v \in V\}$.
    \item \textbf{Adjacency List:} Adjacency List encodes neighboring relationships for each node. The graph is represented as a set of nodes $V$, where each node $i$ has an associated list of neighbors $N_i = \{n_1, n_2, \dots, n_k \}$.
    \item \textbf{Node2Vec Embeddings:} Node2Vec Embeddings are generated using Node2Vec\cite{grover2016node2vec}, which applies random walks followed by Word2Vec to learn node embeddings that capture both local and global structural features. Specifically, for each node $v \in V$, Node2Vec simulates multiple random walks of fixed length $l$, generating sequences $\{(v_1, v_2, \dots, v_l)\}$. A Word2Vec model is then trained on these sequences to learn a low-dimensional embedding $f(v_i) \in \mathbf{R}^d$ for each node. In our experiments, the dimension of Node2Vec embedding is 32.
    \item \textbf{Spectral Embeddings\cite{spectral}:} Spectral embeddings are derived from the eigenvectors of the graph Laplacian matrix. Given an undirected graph $G = (V, E)$ with adjacency matrix $A$ and degree matrix $D$, the graph Laplacian is defined as:
    \[
    L = D - A.
    \]
    The spectral embedding is obtained by computing the first $k$ nontrivial eigenvectors (associated with the smallest non-zero eigenvalues) of $L$. These eigenvectors form a matrix $U \in \mathbf{}{R}^{|V| \times k}$, where each row $U_i$ provides a $k$-dimensional embedding for node $i$. In our experiments, we choose $ k = 8$.
\end{itemize}

\textbf{Coordinate Features:} These are the coordinates of the graph layout produced by the layout algorithm (e.g., neato). Each node $v_i \in V$ is assigned a 2-dimensional coordinate $(x_i, y_i) \in \mathbf{R}^2$, resulting in a position matrix $P \in \mathbf{R}^{|V| \times 2}$. This representation captures the geometric structure of the graph layout.

\begin{table}[h]
\vspace{-0.2cm}
\centering
    \caption{Prompt for few-shot with embeddings evaluation of layout aesthetics. This is added to the prompt in Table~\ref{tab:prompt_zero_shot_image_only}, just before ``Let's think step by step''.}
    \label{tab:embeddings}
    \begin{tabular}{|p{0.45\textwidth}|}
    \hline
    \small{

    ... Below are examples of layouts labeled as "good" or "bad" based on human feedback:
    
    \{examples\}
    
    Now, please evaluate the new graph:

    Graph Structure:
    \{graph structural information\}

    Layout Coordinates to Evaluate:
    \{features\}
    ...
  }\\
    \hline
    \end{tabular}
\vspace{-0.5cm}
\end{table}

\begin{figure}[]
    \centering
    \includegraphics[width=\linewidth]{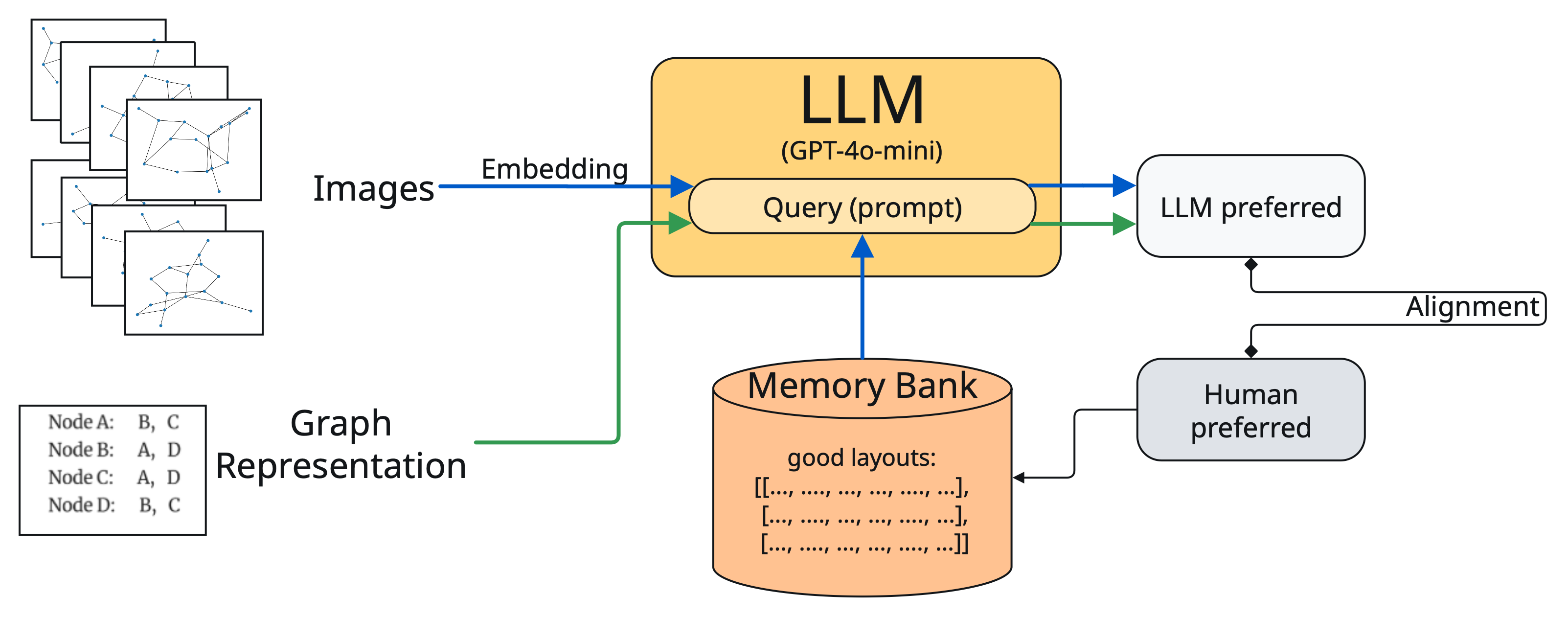}
    \vspace{-0.4cm}
    \caption{Demonstration of memory bank vs graph representation. The green route shows the graph representation, while the blue route shows the image-embedding–based memory bank approach.
    }
    \label{fig:memory_bank_vs}
    \vspace{-0.5cm}
\end{figure}

\subsection{Few-shot with image embedding based memory-bank}

We also explored a memory bank (MB) approach to enhance model decision-making for new layouts by leveraging contextual information from various shot configurations (Figure~\ref{fig:memory_bank_vs}). An MB serves as a repository of embeddings from past examples, allowing the model to draw on similar instances to improve decisions on new layouts. Specifically, we employed embeddings from ResNet-50~\cite{He_2016_CVPR} and DINOv2~\cite{Oquab_2023_DINOv2}. For DINOv2 embeddings, we used the DINOv2-base model, which provides a 768-dimensional representation, in both 1-shot and 5-shot settings. In the 10-shot setting, we used the DINOv2-small model, yielding a 384-dimensional representation to accommodate input size constraints.

For ResNet-50, in the 1-shot and 5-shot scenarios, we fused the outputs of the last four layers and applied pooling to obtain a 768-dimensional representation. In the 10-shot setting, we reduced the pooled representation to 384 dimensions to comply with the input token limitations of GPT-4o-mini, which has a maximum token capacity of 120,000. This adjustment was necessary because the 10-shot configuration could exceed 144,000 tokens, requiring dimensionality reduction to ensure compatibility. The prompts used for both models are provided in Table~\ref{tab:prompt_memory_bank}.

\begin{table}[h]
\vspace{-0.4cm}
\centering
    \caption{Prompt for image-embedding memory bank approach}
    \label{tab:prompt_memory_bank}
    \begin{tabular}{|p{0.45\textwidth}|}
    \hline
    \small{You are an expert in human aesthetics. Below are examples of layouts labeled as "good" or "bad" based on human feedback:
    
    Good Layouts:
    \{good\_layouts\}
    
    Now, evaluate the following layouts and select the one that is most aligned with human aesthetics:
    Features: \{features\}
    
    And the last sentence must start with "In conclusion, I will pick layout X".}\\
    \hline
    \end{tabular}
    \vspace{-0.5cm}
\end{table}

\section{Aligning Vision Models with Human Aesthetic Preference}

One limitation of LLM-based labeling is that only a few examples of human-preferred layouts can be included in the prompt, due to the input token limit of LLMs. In contrast, a traditional vision model can learn human preferences using the tens of thousands of human labels we have collected. Therefore, we develop two vision model (VM)-based classifiers as a complementary approach to LLM-based labeling. Specifically, we frame the task as a supervised learning problem: given eight visualization images, the classifiers perform multi-class classification, aiming to assign the highest logit to the layout preferred by humans.

\subsection{Vision models}

We employ a ResNet-50 model pretrained on ImageNet as the vision backbone. The final fully connected (classification) layer is removed, and the output from the last convolutional block is used as a feature representation. Each input consists of a batch of 8 candidate layout images per graph, where each image is passed independently through the {\em shared} ResNet-50 encoder. The output of each image is a 2048-dimensional feature vector, resulting in an 8×2048 concatenated representation for the eight images.
The concatenated feature vector is fed into a custom Multi-Layer Perceptron (MLP) with the following layer dimensions: $8 \times \text{(embedding \ dim)} \rightarrow 4096 \rightarrow 1024 \rightarrow 8$. The final output is an 8-dimensional vector, where each dimension corresponds to a candidate layout. The model is trained to assign the highest score to the layout that aligns best with the target human aesthetic preference.

In addition to ResNet-50, we explore the use of self-supervised visual representations by incorporating DINOv2\cite{Oquab_2023_DINOv2} as the image encoder. We choose DINOv2 over the newer DINOv3 \cite{siméoni2025dinov3} primarily because we found DINOv2 to be significantly more stable in both training and feature extraction, and DINOv3 does not provide consistently better performance for our task. We experiment with three model sizes: DINOv2-Small, DINOv2-Base, and DINOv2-Large, with corresponding output embedding dimensions of 384, 768, and 1,024, respectively. Similar to the ResNet-50 setup, each of the 8 candidate layout images per graph is processed independently through a {\em shared} DINOv2 encoder. This results in an $8 \times \text{(embedding\ dim)} $ concatenated feature representation for the full candidate set. The resulting concatenated feature vector is then passed through a Multi-Layer Perceptron (MLP) with the same configuration as the one used in the ResNet-50 classifier setup. 

One challenge in our dataset is that humans do not always prefer the same visualization for a given graph. As a baseline model, to stabilize training, we retain only graphs with a single dominant label and use those to train our vision models. This yields 519 training graphs.


\subsection{Data augumentation}
To address the limited size of the training set in the baseline model (519 graphs for training, compared to 1000 graphs for testing) and to improve the model’s generalizability, we apply an augmentation strategy based on randomized permutations of the candidate layouts per graph. 
For each graph $G$, let 
$
X_G = [x_{G,1}, x_{G,2}, \dots, x_{G,8}]
$
denotes the 8 candidate layout images input, and $y_G = [y_{G,1}, y_{G,2}, \dots, y_{G,8}]$
their corresponding one-hot label vectors. Then for each augmentation iteration k, we sample a permutation 
$\pi^{(k)} \in S(8)$,
and construct the augmented sample as
$$
X_G^{(k)} = \bigl[x_{G,\pi^{(k)}(1)},\; x_{G,\pi^{(k)}(2)},\; \dots,\; x_{G,\pi^{(k)}(8)}\bigr],
$$
$$
y_G^{(k)} = \bigl[y_{G,\pi^{(k)}(1)},\; y_{G,\pi^{(k)}(2)},\; \dots,\; y_{G,\pi^{(k)}(8)}\bigr].
$$

Each augmented sample contains the same set of layouts, and the only difference is in the ordering of layouts within the input sequence and the corresponding label. This procedure is repeated $k$ times for each graph to generate the augmented training set of 5190 training samples. We set $k=10$ to balance performance and training efficiency, given that larger $k$ would require more computing time. E.g., our best configuration requires approximately 1/2 day to train on 5K samples due to the complexity of the DINOv2 backbone and fine-tuning.


This strategy offers two benefits: first, it effectively enlarges the training set, helping the model learn more stable patterns despite the smaller number of original training examples. Second, it eliminates positional bias by preventing the model from memorizing which layout tends to appear in which position in the original dataset, instead forcing it to develop true understanding of the visualization, since the same set of 8 images appears in different orders after permutation.

\vspace{-0.3cm}
\subsection{Modeling human diversity with soft multiclass loss}

We would like to use all human labels collected for training (55,828 labels for the 10,000 training graphs), not only those from graphs with unanimous agreement. To do this, we must accommodate the diversity of human preferences. For any given graph and its eight associated visualizations, multiple layouts may be selected by different human labelers. There are three potential ways to handle this.

The first option, which we tested in our baseline and data augmented models, is to only choose the 5.15\% of graphs that all lablers achieved unanimous agreement. While this gives high-quality training data, most of the human labels are wasted, and we had to rely on data augmentation to help improve the training.

Another possible option is majority voting, which selects the visualization favored by the largest number of labelers as the class label. However, this approach still discards valuable labeling information and introduces ambiguity in the case of ties. Therefore, we did not pursue this option.

Unlike those two approaches, a more principled alternative is to treat human preferences as soft multiclass labels rather than a single one-hot target.
For each graph $G$, let the labeler's choices among the eight visualizations be summarized as a probability distribution $\mathbf{q}(G) = [q_1(G), q_2(G), \ldots, q_8(G)]$, where $q_k(G)$ is the fraction of human labelers who selected visualization $k$.
Denote the predicted probability distribution obtained from the softmax layer of the model as
\begin{equation}
\mathbf{p}(G, \theta) = [p_1(G, \theta), p_2(G, \theta), \ldots, p_8(G, \theta)],\label{eqn:prob_dist}
\end{equation}

\noindent where $\theta$ denotes the model parameters.

Then the soft multiclass loss (cross-entropy with soft targets) for a single graph is

$$
\mathcal{L}(G, \theta) = - \sum_{k=1}^{8} q_k(G) \log p_k(G, \theta)
$$

\noindent and the total loss over the dataset (or a batch) is the average loss over the graphs in the dataset (or the batch).
This formulation makes full use of all the labels while preserving the full diversity of human preferences. It allows the model to learn from probabilistic rather than hard labels, and it gracefully handles disagreement and ties among labelers. This approach is also more efficient than treating each label from human \revision{lablers}\removed{annotators} as a separate sample, thereby reducing the training data size from 55,828 to 10,000 and improving training efficiency. 


\subsection{Fine-tuning}
Fine-tuning is a commonly used technique for adapting pretrained visual encoders to downstream tasks. It allows the visual encoder to adapt its representation space from generic image features to the more specialized characteristics of graph layout aesthetics. In this work, we consider two settings for each model (ResNet-50 and DINOv2-Base). In the \emph{frozen} setting, the backbone parameters are kept fixed, and only the MLP classifier on top is trained. In the \emph{fine-tuned} setting, backbone parameters for a few top layers are updated jointly with the MLP. Both settings use the same supervision method introduced in the previous sections (including data augmentation and soft multiclass labels).

\section{Experimental results}

We split the dataset the same way as~\cite{wang_2023_smartgd}, with  10,000 graphs for training, 1000 for testing, and 531 for validation. For LLMs, we use GPT-4o-mini as our main model, selected for its efficiency and visual reasoning capabilities, but we also experimented with other models.

\begin{table}[htbp]
\vspace{-0.2cm}
    \centering
    \caption{Alignment (\%) of LLM with human labels. Note that 10 shots for Node2Vec experiments are missing due to token length limits. Rows marked with * do not use coordinates, and the rest use coordinates.
} 
    \label{tab:LLM_with_individual_human}
    \sisetup{
    table-number-alignment = center,
    round-mode            = places,
    round-precision       = 2
    }
        
    \begin{tabular}{
      l
      S[table-format=2.2]
      S[table-format=2.2]
      S[table-format=2.2]
      S[table-format=2.2]
    }
    \toprule
    \textbf{Method} & \textbf{0‑Shot} & \textbf{1‑Shot} & \textbf{5‑Shot} \\
    \midrule
    Image Only*    & 19.70 &  19.52 &  18.79 \\
    Edgelist      & 16.76 & 16.21 & 14.35 \\
    Adjacency     & 16.37 & 16.92 & 15.77 \\
    Node2Vec   & 17.67 & 15.01 & {---} \\
    Spectral      & 17.21 & 15.71 & 14.54 \\
    MB ResNet50*   & 10.45 & 11.37 & 14.45 \\
    MB DINOv2*     & 10.43 & \textbf{33.11} & 30.25 \\
    \midrule
    {Human (Ideal)} & \multicolumn{3}{c}{{38.34}} \\
    \bottomrule
    \end{tabular}
        \vspace{-0.5cm}

\end{table}


\subsection{LLM performance}

The results of different prompting strategy are given in Table~\ref{tab:LLM_with_individual_human}. As we can see, alignment with humans is better with image-only vs using a combination of graph structure information and coordinate features. However, among the prompting strategies, the prompt that uses an image-embedding–based memory bank with the DINOv2 model achieves the best performance, reaching an alignment of 33.11\% in the one-shot setting. The improved performance of MB DINOv2, which relies solely on visual representations without explicit coordinate information, suggests that rich pretrained visual embeddings can effectively capture aspects of human aesthetic preference that structured graph encodings may miss. Nevertheless, a small gap remains relative to ideal human–human agreement (33.11\% vs 38.34\%). Even with strong visual encoders, LLMs still only partially approximate the nuanced aesthetic criteria humans apply when evaluating graph layouts.

Another interesting finding is that for image-only prompts and for structural-plus-coordinate prompts (Edgelist, Adjacency, Node2Vec, and Spectral), performance drops as k-shot examples are provided. This suggests that the LLM may already possess some understanding of human aesthetic preferences, and that limited shot-based prompting can sometimes confuse the model. A few examples may reinforce surface-level patterns rather than improving the model’s true understanding of graph aesthetics. In contrast, for image-embedding–based memory bank models, where the LLM does not have direct access to the image or node coordinates, few-shot examples are important, as they provide labeled guidance on how to effectively use the visual embeddings.

To further evaluate the generality of the DINOv2 memory-bank 1-shot strategy across different large language models, we conducted the same experiment using several frontier LLMs. As summarized in Table \ref{tab:llm_comparison}, while most models show non-trivial alignment with human preferences, their performance varies notably. Qwen3 achieved only 11.77\%, Claude Sonet 4.5 12.70\%, and Gemini 2.5 Flash 14.81\%. GPT-5 performed substantially better at 27.80\%, but GPT-4o mini remains the strongest model, attaining 33.11\% alignment, approaching human–human agreement (38.34\%). These findings highlight that both the visual reasoning ability of the LLM models, and aesthetic alignment training via 1-shot, critically affect performance under identical DINOv2 embeddings.

\begin{table}[h]
\vspace{-0.2cm}
\centering
\caption{\revision{LLM}\removed{VM}-human alignment (\%) for DINOv2 Memory Bank 1-shot across different LLMs.}
\label{tab:llm_comparison}
\begin{tabular}{lc}
\toprule
\textbf{Model} & \textbf{Alignment (\%)} \\
\midrule
Qwen3 (235b-a22b-2507) & 11.77 \\
Claude Sonet 4.5 & 12.70 \\
Gemini 2.5 Flash & 14.81 \\
GPT-5 & 27.80 \\
GPT-4o mini & \textbf{33.11} \\
\bottomrule
\end{tabular}
\vspace{-0.3cm}
\end{table}


\subsection{VM performace}

We also evaluate the ability of VM on our graph labeling task (Table~\ref{tab:VM_results}).
Across all configurations, DINOv2 consistently outperforms ResNet-50, indicating its stronger ability to approximate human layout preferences. \revision{A paired-sample t-test confirmed that this difference is statistically significant.} Table \ref{tab:VM_results} ummarizes the results under different training settings for the VM models. Among all variants, the soft-multiclass strategy with fine-tuning using DINOv2 achieves the highest VM-human alignment score of 36.81\%, very close to the human–human alignment of 38.34\%. This is also slightly better than LLM-human alignment (33.11\%), showing that VM holds a modest advantage over LLMs for this task, likely due to the fact that it is possible to ``teach'' VMs the human preferences via tens of thousands of examples, vs a limited few shots for LLM. 

Compared with the baseline that uses only the labels for the 5.15\% of graphs with unanimous human agreement, both of our proposed strategies, data augmentation and soft-multiclass training, lead to clear improvements across models. Augmentation provides moderate gains for both ResNet and DINOv2 in the frozen-feature-extractor setting, and for ResNet in the fine-tuned-backbone setting. Soft-multiclass supervision yields a substantially larger boost, especially for ResNet (from 16.82\% to 30.41\%, and from 17.36\% to 35.03\%) in both settings. When fine-tuning is enabled (top 2 layers), most configurations outperform their frozen counterpart. This highlights the potential benefits of adapting the backbone.

\begin{table}[t]
\centering
\small
\setlength{\tabcolsep}{6pt}
\caption{VM-human alignment (\%) under different training strategies for ResNet and DINOv2}
\label{tab:VM_results}
\vspace{0.5em}
\begin{tabular}{lcc}
\toprule
\textbf{Method} & \textbf{ResNet} & \textbf{DINOv2} \\
\midrule
\multicolumn{3}{l}{\textit{Frozen feature extractor}} \\
Baseline                    & 16.82 & 27.82 \\
Augment (10×)               & 18.66 & 32.82 \\
Soft Multiclass             & 30.41 & 32.80 \\
\midrule
\multicolumn{3}{l}{\textit{Fine-tuned backbone}} \\
Tune + Baseline             & 17.36 & 33.07 \\
Tune + Augment              & 18.37 & 33.41 \\
Tune + Soft Multiclass      & 35.03 & \textbf{36.81} \\
\bottomrule
\end{tabular}
\end{table}

To isolate the effects of model size and the number of trainable layers in the DINOv2 backbone, we conducted two ablation studies based on the best-performing configuration. Table~\ref{tab:vm_ablation_size} reports results obtained using different DINOv2 backbones. Interestingly, the DINOv2-large model does not outperform the DINOv2-base variant, suggesting that simply increasing backbone capacity does not necessarily improve alignment with human preferences for this task. Table~\ref{tab:vm_ablation_layer_size} shows the effect of unfreezing 1, 2, 3, or 4 layers of DINOv2-base. \removed{Unfreezing 2 layers yields the best alignment, with 4 layers a close second.}
\revision{We conducted paired-sample t-tests across all configurations, and the results indicate that model size does not make a significant difference in performance.}


\begin{table}[t]
\vspace{-0.2cm}
\centering
\small
\setlength{\tabcolsep}{8pt}
\caption{Ablation study of VM-human alignment with three DINOv2 models of different sizes, under fine-tuning setting (2 layers)}\label{tab:vm_ablation_size}
\vspace{0.5em}
\begin{tabular}{lccc}
\toprule
\textbf{Model} & \textbf{Size} & \textbf{Embedding Dim} & \textbf{Alignment (\%)} \\
\midrule
DINOv2-Small & 21M & 384  & 35.05 \\
DINOv2-Base  & 86M  & 768  & \textbf{36.81} \\
DINOv2-Large & 300M & 1024  & 35.90 \\
\bottomrule
\end{tabular}
\vspace{-0.5cm}
\end{table}

\begin{table}[t]
\vspace{-0.2cm}
\centering
\small
\setlength{\tabcolsep}{8pt}
\caption{Ablation study for number of layers to fine-tune}\label{tab:vm_ablation_layer_size}
\vspace{0.5em}
\begin{tabular}{lcc}
\toprule
\textbf{Model} & \textbf{Num. of fine-tune layers} & \textbf{Alignment (\%)} \\
\midrule
DINOv2-Base  & 1  & 32.10 \\
DINOv2-Base  & 2  & \textbf{36.80} \\
DINOv2-Base  & 3  & 34.76 \\
DINOv2-Base  & 4  & 36.04 \\
\bottomrule
\end{tabular}
\vspace{-0.5cm}
\end{table}

\subsection{Further Improvements in AI–Human Alignment}

As we have seen thus far, VM-human alignment is 36.81\%, close to the human-human alignment of 38.34\%. There is a larger gap between LLM-human alignment (33.11\%) and human-human alignment. We are interested in whether the alignment improves if we only use higher-confidence LLM or VM labels. For the LLM, confidence is defined as the log-probability assigned to the predicted layout label token, exponentiated to obtain the probability. For the VM model, given the model-predicted probability distribution for a graph (Equation~\ref{eqn:prob_dist}), we set confidence to 
$\max_{i\in\{1,2,\ldots,8\}} p_i(G,\theta)$.

\begin{figure}[htbp!]
        \includegraphics[width=0.5\textwidth]{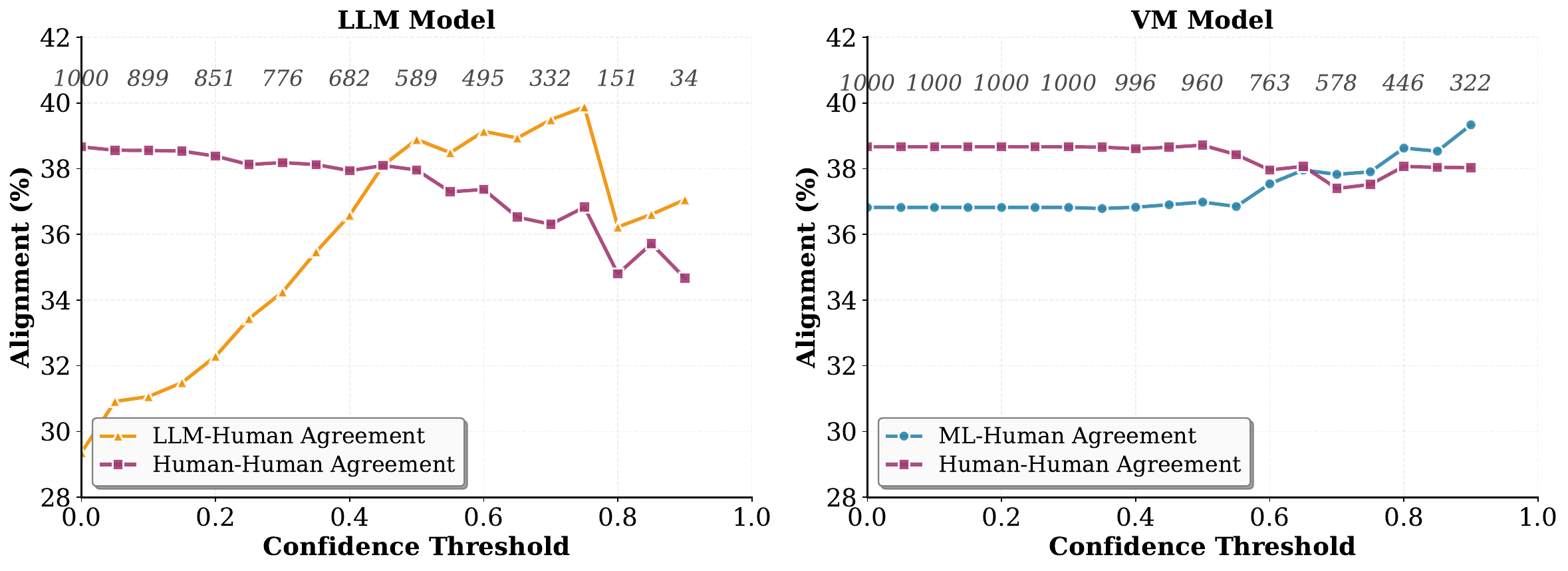} 
    \caption{Left: Human-human and LLM-human alignment as a function of VM confidence. Right: Human-human and VM-human alignment as a function of VM confidence. The number of test samples remaining after threadholding is given at the top of each figure.
    \label{fig:confidence}}
\vspace{-0.9cm}
\end{figure}

Figure~\ref{fig:confidence} shows that both LLM-human and VM-human alignment improve when imposing a modest confidence threshold. The models are not uniformly strong across all graphs, but their high-confidence predictions tend to align more closely with human preferences. An interesting observation from the LLM confidence-threshold analysis (Figure~\ref{fig:confidence}, left) is that the LLM (yellow line) exhibits a rapid rise, even surpassing human-human alignment (red line). For example, when setting the threshold to 0.45, LLM is indistinguishable from a human labeler, and can still provide labels for around 65\% of the graphs (numbers at the top of the chart).


On the other hand, VM-human alignment (blue line in Figure~\ref{fig:confidence}, right) starts very close to human-human alignment (red line) and continues to improve as the confidence threshold increases, eventually surpassing the average human-human alignment at higher thresholds. For example, at threshold 0.6, VM is on par with the average human labeler, and can provide labels for around 76\% of the graphs. It is also interesting that human-human alignment is generally flat, but dips slightly to around 37\% for the graphs on which the VM has high confidence. For the graphs on which the LLM is highly confident, human-human alignment drops more substantially, to around 34\%. This is likely because the VM is trained with 50K human labels and therefore aligns better with average human preferences, whereas the 1-shot LLM relies largely on innate knowledge derived from pretraining, and therefore behaves complementary to the human labelers.

\subsection{Understanding the AI labelers}

We further explore the distribution of preferred layouts across humans, VM, and LLM in Table~\ref{tab:vm_llm_human_layout_choices}. Kamada–Kawai dominates both human choices (42.54\%) and VM predictions (49.50\%), suggesting that VM captures the global preference trend well. It also dominates LLM choices, but accounts for an overwhelming 70.90\%, even though images are randomly permuted before querying the LLM. Overall, we found that the distribution of the preferred layout algorithms for humans and VMs are quite similar.

\begin{figure}[htbp!]
\begin{center}

        \includegraphics[width=0.4\textwidth]{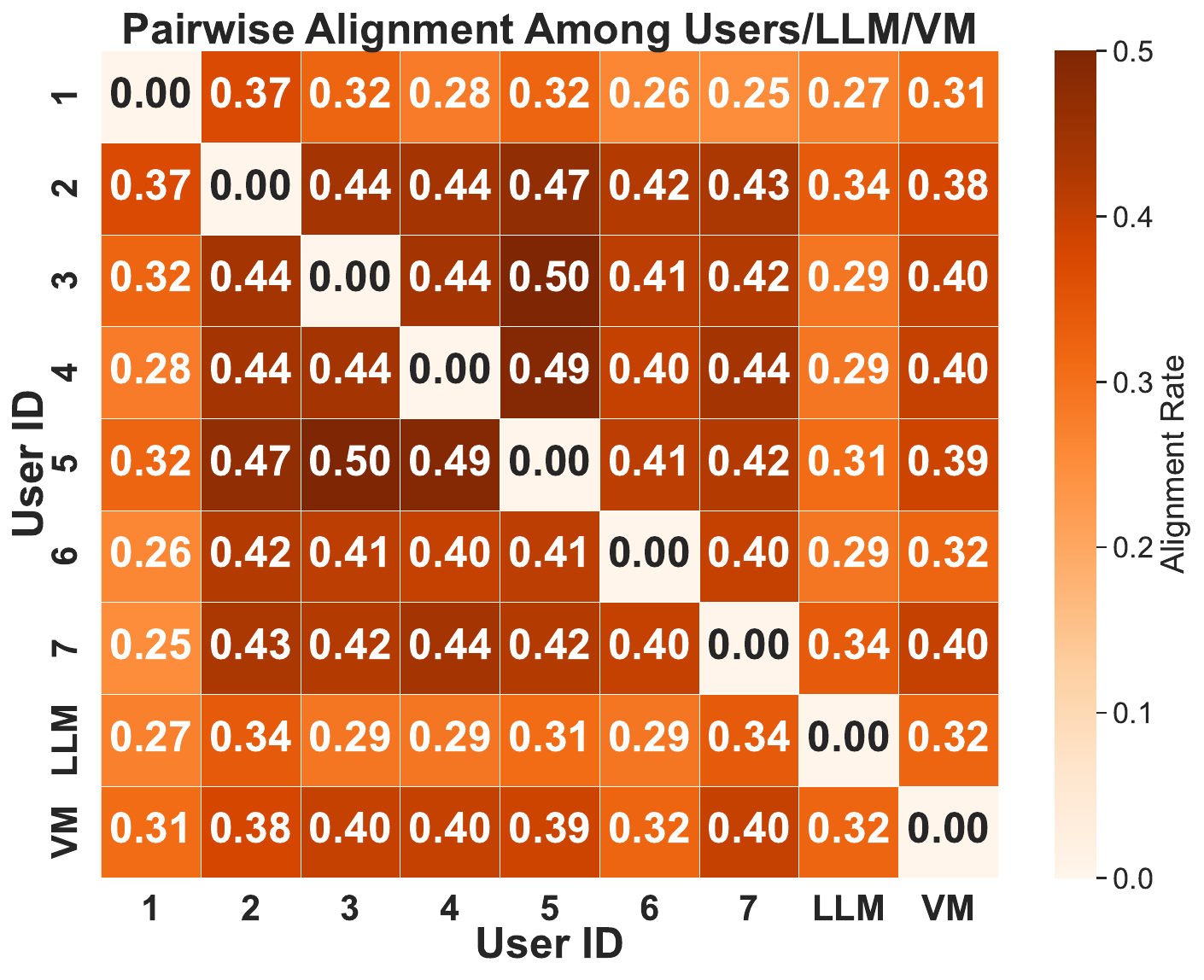} 
\vspace{-0.4cm}
    \caption{Human-human, LLM-human and VM-human alignment visualization. Here we pick the seven most prolific human labelers.
    \label{fig:human_human_llm_vm}}
\end{center}
\vspace{-0.7cm}
\end{figure}

The analyses above show that VM–human alignment is very close to human–human alignment. LLM–human alignment can also approach the human–human level when only high-confidence labels are considered. To examine whether aggregated AI–human alignment masks variance across real human labelers, we visualized the alignment patterns in Figure~\ref{fig:human_human_llm_vm}. For the top seven most prolific human labelers, we compare human–human, LLM–human, and VM–human pairwise alignment. Human labelers 3 and 5 have the closest choices (50\%), but no two individuals agree perfectly. The VM (last column/row) behaves like a typical human annotator, with no discernible differences. The LLM (second-to-last column/row) shows slightly lighter color, indicating somewhat lower alignment with the human labelers.


\subsection{Examples}

\begin{figure}[h]
\centering
\begin{tabular}{c}
    \includegraphics[width=0.95\linewidth]{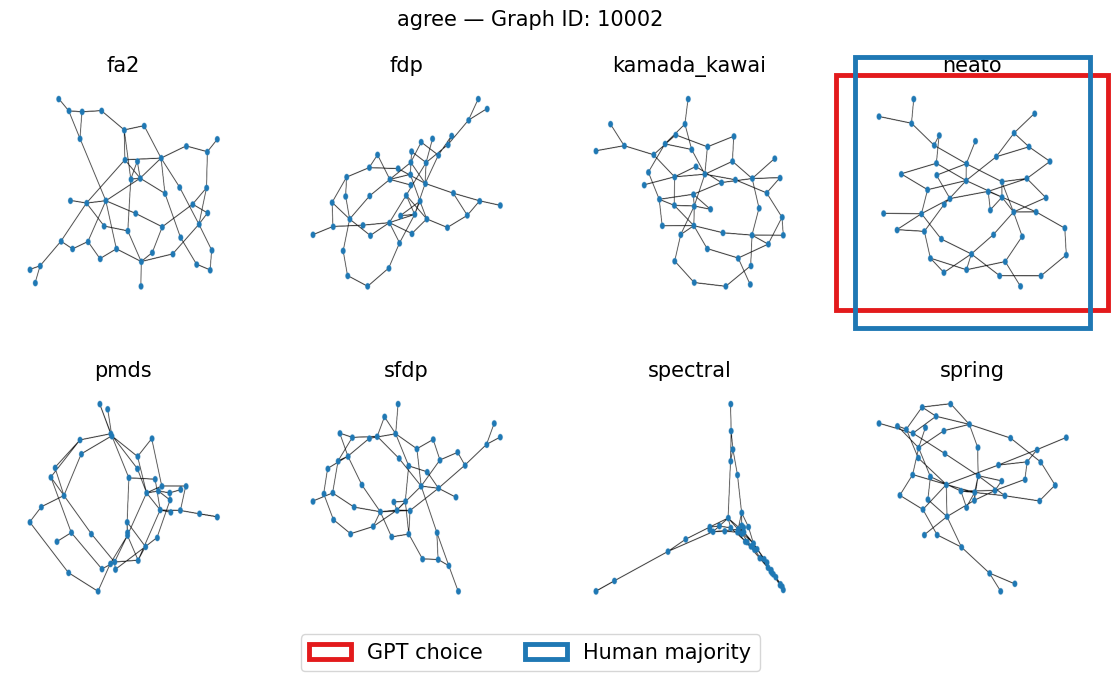} \\[0.3em]

    \begin{minipage}{0.95\linewidth}
        \centering
        (a) Example where LLM and human \textbf{agree} on layout
    \end{minipage}
    \\[1.0em]

    \includegraphics[width=0.95\linewidth]{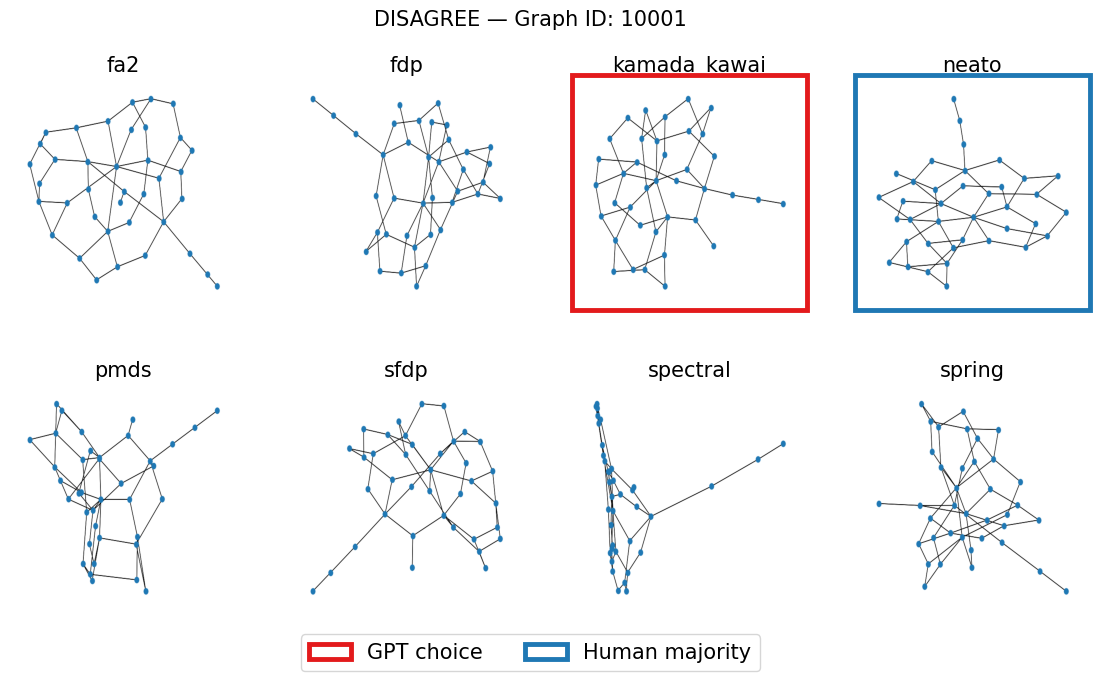} \\[0.3em]

    \begin{minipage}{0.95\linewidth}
        \centering
        (b) Example where LLM and human \textbf{disagree} on layout
    \end{minipage}
\end{tabular}
\vspace{-0.2cm}

\caption{Examples illustrating agreement and disagreement between LLM and human preferences.}
\label{fig:gpt_human_agreement}
\vspace{-0.7cm}
\end{figure}

\begin{figure}[h]
\centering
\begin{tabular}{c}
    \includegraphics[width=0.95\linewidth]{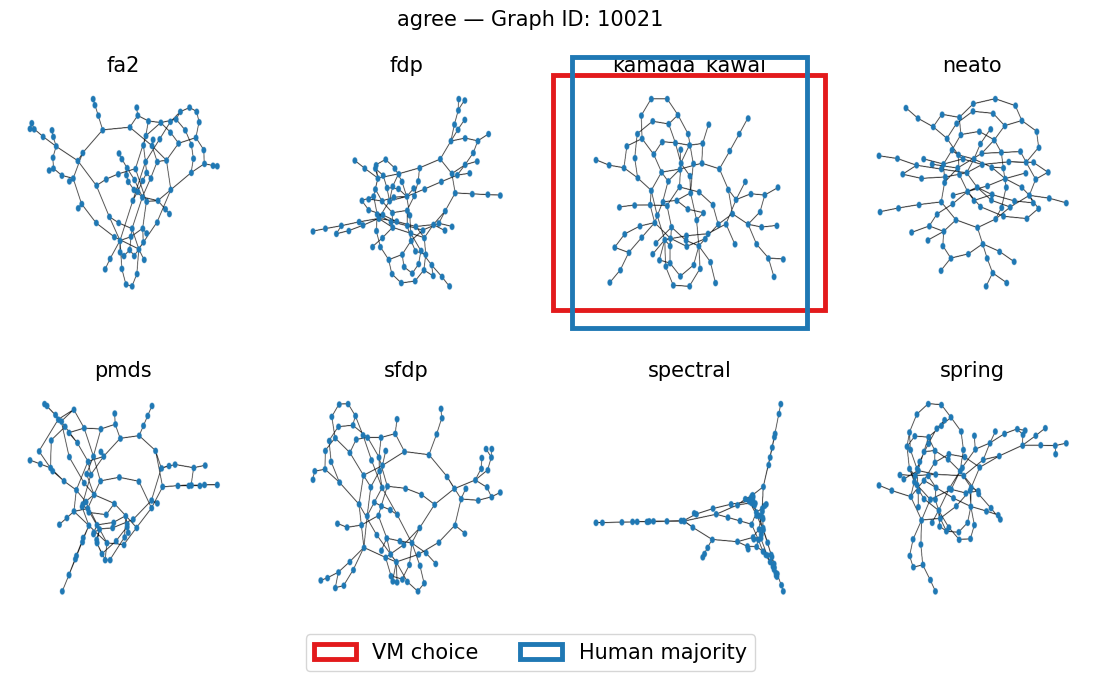} \\[0.3em]

    \begin{minipage}{0.95\linewidth}
        \centering
        (a) Example where VM and human \textbf{agree} on layout
    \end{minipage}
    \\[1.0em]

    \includegraphics[width=0.95\linewidth]{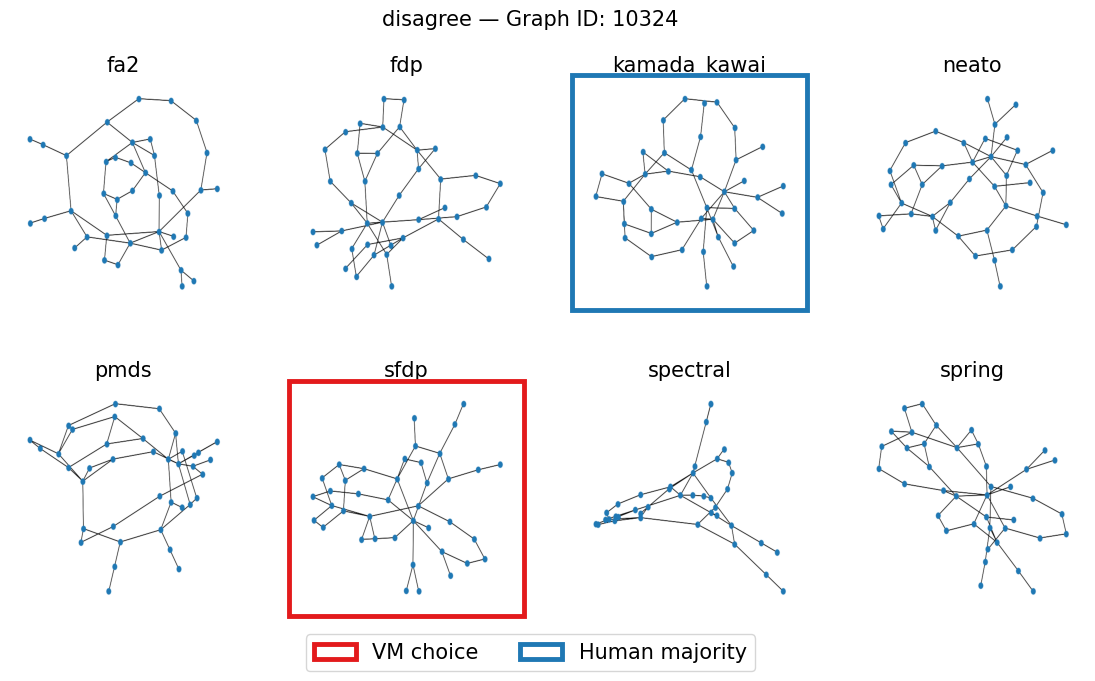} \\[0.3em]

    \begin{minipage}{0.95\linewidth}
        \centering
        (b) Example where VM and human \textbf{disagree} on layout
    \end{minipage}
\end{tabular}
\vspace{-0.2cm}

\caption{Examples illustrating agreement and disagreement between VM and human preferences.}
\label{fig:vm_human_agree_examples}
\vspace{-0.7cm}
\end{figure}

To further illustrate how VMs and LLMs behave in comparison to human \revision{lablers}\removed{annotators}, we present several examples in this section. Figure~\ref{fig:gpt_human_agreement} shows cases where the LLM (a) agrees and (b) disagrees with human-preferred layouts. In the disagreement case (b), the LLM’s choice looks quite similar to the human choice, though not exactly the same. Figure~\ref{fig:vm_human_agree_examples} shows analogous cases for the VM, where it (a) agrees and (b) disagrees with human-preferred layouts. In the disagreement case (b), the VM’s choice also appears reasonable. These examples demonstrate that perfect alignment is unrealistic and that diverse, yet still defensible, choices are to be expected when modeling human aesthetic preferences.

\begin{table}[h]
\vspace{-0.3cm}
\setlength{\tabcolsep}{2.9pt}
\centering
\caption{Choice of layout algorithms by LLM (ChatGPT-40-mini), VM (DINOv2 Tune $+$ Soft Multiclass) and Human. Note that the ``Human Count'' column has a total of 5557 labels, because each of the 1000 test graphs has, on average, 5.56 labels by different human labelers.}
\begin{tabular}{c|cc|cc|cc}
\hline
\textbf{Layout} &
\makecell{\textbf{LLM}\\ \textbf{Count}} &
\makecell{\textbf{LLM}\\ \textbf{\%}} &
\makecell{\textbf{VM}\\ \textbf{Count}} &
\makecell{\textbf{VM}\\ \textbf{\%}} &
\makecell{\textbf{Human}\\ \textbf{Count}} &
\makecell{\textbf{Human}\\ \textbf{\%}} \\
\hline
fa2           & 31  & 3.10  & 5   & 0.50  & 269  & 4.84 \\
fdp           & 193 & 19.30 & 36  & 3.60  & 448  & 8.06 \\
Kamada-Kawai & \textbf{709} & \textbf{70.90} & \textbf{495} & \textbf{49.50} & \textbf{2364} & \textbf{42.54} \\
neato         & 22  & 2.20  & 436 & 43.60 & 1935 & 34.82 \\
pmds          & 9   & 0.90  & 4   & 0.40  & 72   & 1.30 \\
sfdp          & 8   & 0.80  & 15  & 1.50  & 292  & 5.25 \\
spectral      & 3   & 0.30  & 0   & 0.00  & 6    & 0.11 \\
spring        & 25  & 2.50  & 9   & 0.90  & 171  & 3.08 \\
\hline
\end{tabular}
\label{tab:vm_llm_human_layout_choices}
\vspace{-0.2cm}
\end{table}

\subsection{Findings for RQ2 and RQ3} 

Based on the above results, we can conclude that 

\begin{itemize}

\item Regarding \textbf{RQ2}, with careful prompt engineering using a memory bank and self-supervised vision-encoder features, LLM–human alignment can be improved from 19.7\% (0-shot image-only) to 31.11\%. While a gap to human–human alignment (38.34\%) remains, this gap can be closed by applying a confidence threshold, retaining approximately 65\% of the labels.

\item Regarding \textbf{RQ3}, the fined-tuned DINOv2 VM with soft multiclass loss function performs very close to an average human labeler (36.81\% vs.\ 38.34\%). When thresholding the VM labels, the VM can match or surpass the average human labeler, while still retaining around 76\% of the labels.

\end{itemize}

Overall, we believe these results indicate that AI can perform at a level that makes it an effective proxy for human labelers.

\section{Limitations}

We collected human-labeled data using a within-subject design where participants selected their preferred visualization from eight options, unlike prior pairwise studies~\cite{Cai2021PacificVis,Wu2025GNNGraph}. \revision{At the cost of reduced interpretability and 
less fine-grained relative preference information compared to 
pairwise or ranking-based approaches,} this design allows efficient and simultaneous comparison of eight layout algorithms, and builds on SmartGD~\cite{wang_2023_smartgd}, which used a similar setup but selected the best layout based on metrics. While efficient for 11K graphs, this format also guides LLM and VM labeling. A limitation is that the trained VM cannot directly serve as a GAN discriminator~\cite{sgan}, which typically operates on graph pairs~\cite{wang_2023_smartgd}. Instead, VM or LLM models act as human proxies to generate large-scale preference labels, which can then train a GAN as in SmartGD. \revision{In addition, the study's long three-year duration and participant heterogeneity may introduce confounds, such as evolving aesthetic trends. The participant pool, while typical of graph visualization users, is also not fully representative of the broader population, warranting caution in generalizing our findings.}

\revision{Our primary LLM findings are based on GPT-4o-mini, and the substantially varying alignment scores across models in Table~\ref{tab:llm_comparison} suggest that the effectiveness of our methodology is sensitive to the choice of LLM, warranting broader validation in future work.}
Both the best LLM and VM models rely on rendered images rather than underlying layout data, which aligns with the goal of producing human-readable visualizations. However, this requires generating images before using the models as human-proxy labelers. We plan to investigate whether layout information alone could achieve the same goal, bypassing rendering.

\section{Conclusion}

In this paper, we investigated whether large language models (LLMs) and vision models (VMs) can serve as effective proxies for human preferences in network visualization. We conducted a user study with 27 participants, where each participant selected their preferred visualization from eight alternatives of the same graph. Analysis of the 64K human-preference labels showed that human preferences are both structured and diverse, with an average human–human alignment of 38.34\%. We found that an LLM (ChatGPT-4o-mini), equipped with a DINOv2-based memory bank, achieves 33.11\% alignment with human labelers and can match human-level performance when lower-confidence labels are filtered out. We also showed that a VM (DINOv2), carefully trained to reflect diverse human tastes, can learn to model human preferences at a level comparable to an average human labeler. Overall, our findings indicate that AI, especially VMs, can feasibly serve as scalable proxies for human labelers. \revision{For future work, while we focus on modeling average human preferences in this paper, learning personalized aesthetic preferences that account for individual diversity remains an interesting direction; in addition, we would like to extend our work to non-straightline drawing styles (e.g., orthogonal, hierarchical).}

\clearpage
\bibliography{ref}

@string{lncs = "LNCS"}

@string{gd04 = "Proc. 12th Intl. Symp. Graph Drawing (GD '04)"}

@string{congress = "Congressus Numerantium"}

@article{wang2020deepdrawing,
  title   = {DeepDrawing: A Deep Learning Approach to Graph Drawing},
  author  = {Wang, Yong and Jin, Zhihua and Wang, Qianwen and Cui, Weiwei and Ma, Tengfei and Qu, Huamin},
  journal = {IEEE Transactions on Visualization and Computer Graphics},
  volume  = {26},
  number  = {1},
  pages   = {676--686},
  year    = {2020},
  doi     = {10.1109/TVCG.2019.2934798}
}

@inproceedings{He_2016_CVPR,
  author    = {Kaiming He and
               Xiangyu Zhang and
               Shaoqing Ren and
               Jian Sun},
  title     = {Deep Residual Learning for Image Recognition},
  booktitle = {Proceedings of the IEEE Conference on Computer Vision and Pattern Recognition (CVPR)},
  year      = {2016},
  pages     = {770--778},
  doi       = {10.1109/CVPR.2016.90},
  url       = {https://doi.org/10.1109/CVPR.2016.90},
  publisher = {IEEE},
  bibsource = {DBLP, https://dblp.org}
}

@article{Oquab_2023_DINOv2,
  author    = {Maxime Oquab and
               Timothée Darcet and
               Théo Moutakanni and
               Huy Vo and
               Marc Szafraniec and
               Vasil Khalidov and
               Pierre Fernandez and
               Daniel Haziza and
               Francisco Massa and
               Alaaeldin El-Nouby and
               Mahmoud Assran and
               Nicolas Ballas and
               Wojciech Galuba and
               Russell Howes and
               Po-Yao Huang and
               Shang-Wen Li and
               Ishan Misra and
               Michael Rabbat and
               Vasu Sharma and
               Gabriel Synnaeve and
               Hu Xu and
               Hervé Jégou and
               Julien Mairal and
               Patrick Labatut and
               Armand Joulin and
               Piotr Bojanowski},
  title     = {DINOv2: Learning Robust Visual Features without Supervision},
  journal   = {arXiv preprint arXiv:2304.07193},
  year      = {2023},
  url       = {https://arxiv.org/abs/2304.07193},
  bibsource = {DBLP, https://dblp.org}
}

@article{kamada_kawai_1989, 
title={An algorithm for drawing general undirected graphs}, 
volume={31}, DOI={10.1016/0020-0190(89)90102-6}, 
number={1}, 
journal={Information Processing Letters}, author={Kamada, Tomihisa and Kawai, Satoru}, year={1989}, pages={7–15}}

@article{pmds,
author = {Brandes, Ulrik and Pich, Christian},
year = {2006},
month = {09},
pages = {},
title = {Eigensolver Methods for Progressive Multidimensional Scaling of Large Data},
volume = {4372},
isbn = {978-3-540-70903-9},
journal = {LNCS},
doi = {10.1007/978-3-540-70904-6_6}
}

@article{sfdp,
author = {Hu, Yifan},
year = {2005},
month = {01},
pages = {37-71},
title = {Efficient and High Quality Force-Directed Graph Drawing},
volume = {10},
journal = {Mathematica Journal}
}

@article{fdp,
  title={Graph drawing by force-directed placement},
  author={Fruchterman, Thomas MJ and Reingold, Edward M},
  journal={Software: Practice and experience},
  volume={21},
  number={11},
  pages={1129--1164},
  year={1991},
  publisher={Wiley Online Library}
}

@article{spring,
  title={Graph drawing by force-directed placement},
  author={Fruchterman, Thomas MJ and Reingold, Edward M},
  journal={Software: Practice and experience},
  volume={21},
  number={11},
  pages={1129--1164},
  year={1991},
  publisher={Wiley Online Library}
}

@inproceedings{spectral,
  title={On Spectral Graph Drawing},
  author={Yehuda Koren},
  booktitle={COCOON},
  year={2003}
}

@article{fa2,
author = {Jacomy, Mathieu and Venturini, Tommaso and Heymann, Sebastien and Bastian, Mathieu},
year = {2014},
month = {06},
pages = {e98679},
title = {ForceAtlas2, a Continuous Graph Layout Algorithm for Handy Network Visualization Designed for the Gephi Software},
volume = {9},
journal = {PloS one},
doi = {10.1371/journal.pone.0098679}
}

@inproceedings{neato,
  title={Graph drawing by stress majorization},
  author={Gansner, Emden R and Koren, Yehuda and North, Stephen},
  booktitle={Proc. Springer International Symposium on Graph Drawing},
  pages={239--250},
  year={2004}
}

@article{Fruchterman_Reigold_1991,
  author =      "T. M. J. Fruchterman and E. M. Reingold",
  title =       "Graph drawing by force directed placement",
  journal =     "Software - Practice and Experience",
  volume =      "21",
  year =        "1991",
  pages =       "1129-1164",
}

@article{Eades_1984,
  author =      "P. Eades",
  title =       "A heuristic for graph drawing",
  journal =     congress,
  volume =      "42",
  year =        "1984",
  pages =       "149-160",
}

@inproceedings{Hachul_Junger_2004,
  author = {S. Hachul and M. J\"unger},
  title = {Drawing large graphs with a potential field based multilevel algorithm},
  booktitle   = gd04,
  series = lncs,
  volume = {3383},
  year = {2004},
  pages = 	 {285-295},
  publisher   = {Springer},
}

@article{rome_graphs,
    author = "G. Di Battista and A. Garg and G. Liotta and R. Tamassia and E. Tassinari and F. Vargiu ",
    title = "An Experimental Comparison of Four Graph Drawing Algorithms",
    journal = "CGTA: Computational Geometry: Theory and Applications",
    volume = "7",
    pages = "303-325",
    year = "1997"
}

@article{Purchase_2011_gdaesthetics,
  author = {Purchase, Helen C.},
  ee = {http://dx.doi.org/10.1016/S0953-5438(00)00032-1},
  journal = {Interacting with Computers},
  number = 2,
  pages = {147-162},
  title = {Effective information visualisation: a study of graph drawing aesthetics and algorithms.},
  url = {http://dblp.uni-trier.de/db/journals/iwc/iwc13.html\#Purchase00},
  volume = 13,
  year = 2000
}

@inproceedings{sgan,
  title={Generative Adversarial Nets},
  author={Ian J. Goodfellow and Jean Pouget-Abadie and Mehdi Mirza and Bing Xu and David Warde-Farley and Sherjil Ozair and Aaron C. Courville and Yoshua Bengio},
  booktitle={Proc. NeurIPS},
  year={2014}
}

@article{Argyriou,
author = {Argyriou, Evmorfia and Bekos, Michael and Symvonis, Antonios},
year = {2010},
month = {09},
pages = {},
title = {Maximizing the Total Resolution of Graphs},
volume = {6502},
journal = {Proc. of GD 2010}
}

@article{kobourov_2013, 
title={Force-Directed Drawing Algorithms}, 
DOI={10.1201/b15385-15}, 
journal={Handbook of Graph Drawing and Visualization}, 
author={Kobourov, Stephen G},
year={2013}, 
pages={383–408}}

@article{sgd2,
  title={Multicriteria Scalable Graph Drawing via Stochastic Gradient Descent, $({SGD})^2$},
  author={Ahmed, Reyan and De Luca, Felice and Devkota, Sabin and Kobourov, Stephen and Li, Mingwei},
  journal={IEEE Transactions on Visualization and Computer Graphics},
  volume={28},
  number={6},
  pages={2388--2399},
  year={2022},
  publisher={IEEE}
}

@article{xing-heuristic,
author = {Radermacher, Marcel and Reichard, Klara and Rutter, Ignaz and Wagner, Dorothea},
title = {Geometric Heuristics for Rectilinear Crossing Minimization},
year = {2019},
issue_date = {2019},
publisher = {Association for Computing Machinery},
address = {New York, NY, USA},
volume = {24},
issn = {1084-6654},
journal = {ACM J. Exp. Algorithmics},
articleno = {1.12},
numpages = {21}
}

@ARTICLE{Tim-user-study,  author={Dwyer, Tim and Lee, Bongshin and Fisher, Danyel and Quinn, Kori Inkpen and Isenberg, Petra and Robertson, George and North, Chris},  journal={IEEE Transactions on Visualization and Computer Graphics},   title={A Comparison of User-Generated and Automatic Graph Layouts},   year={2009},  volume={15},  number={6},  pages={961-968},  doi={10.1109/TVCG.2009.109}}

@article{Bekos-xangle,
  title={A heuristic approach towards drawings of graphs with high crossing resolution},
  author={Bekos, Michael A and F{\"o}rster, Henry and Geckeler, Christian and Holl{\"a}nder, Lukas and Kaufmann, Michael and Spallek, Amad{\"a}us M and Splett, Jan},
  journal={The Computer Journal},
  volume={64},
  number={1},
  pages={7--26},
  year={2021},
  publisher={Oxford University Press}
}

@article{kwon-ma-2020,
  title={A deep generative model for graph layout},
  author={Kwon, Oh-Hyun and Ma, Kwan-Liu},
  journal={IEEE Transactions on Visualization and Computer Graphics},
  volume={26},
  number={1},
  pages={665--675},
  year={2019},
  publisher={IEEE}
}

@article{deep-drawing,
  title={Deepdrawing: A deep learning approach to graph drawing},
  author={Wang, Yong and Jin, Zhihua and Wang, Qianwen and Cui, Weiwei and Ma, Tengfei and Qu, Huamin},
  journal={IEEE Transactions on Visualization and Computer Graphics},
  volume={26},
  number={1},
  pages={676--686},
  year={2019},
  publisher={IEEE}
}

@article{deepgd,
author={Wang, Xiaoqi and Yen, Kevin and Hu, Yifan and Shen, Han-Wei},
journal={IEEE Computer Graphics and Applications},
title={Deep{GD}: A Deep Learning Framework for Graph Drawing Using {GNN}},
year={2021},
volume={41},
number={5},
pages={32-44},
doi={10.1109/MCG.2021.3093908}
}

@article{haleem-huamin,
author = {Haleem, Hammad and Wang, Yong and Puri, Abishek and Wadhwa, Sahil and Qu, Huamin},
year = {2019},
month = {07},
pages = {40-53},
title = {Evaluating the Readability of Force Directed Graph Layouts: A Deep Learning Approach},
volume = {39},
journal = {IEEE Computer Graphics and Applications},
doi = {10.1109/MCG.2018.2881501}
}

@inproceedings{purchase1,
  title={Which aesthetic has the greatest effect on human understanding?},
  author={Purchase, Helen},
  booktitle={Proc. Springer International Symposium on Graph Drawing},
  pages={248--261},
  year={1997}
}

@inproceedings{dnn,
  title={Deep Neural Network for DrawiNg Networks, ${(DNN)}^2$},
  author={Giovannangeli, Loann and Lalanne, Frederic and Auber, David and Giot, Romain and Bourqui, Romain},
  booktitle={Proc. Springer International Symposium on Graph Drawing and Network Visualization},
  pages={375--390},
  year={2021}
}

@article{gnn-gd,
  title={Graph neural networks for graph drawing},
  author={Tiezzi, Matteo and Ciravegna, Gabriele and Gori, Marco},
  journal={IEEE Transactions on Neural Networks and Learning Systems},
  year={2022},
  publisher={IEEE}
}

@inproceedings{spx,
  title={Stress-plus-x ({SPX}) graph layout},
  author={Devkota, Sabin and Ahmed, Reyan and De Luca, Felice and Isaacs, Katherine E and Kobourov, Stephen},
  booktitle={Prof. Springer International Symposium on Graph Drawing and Network Visualization},
  pages={291--304},
  year={2019}
}

@article{didimo,
author = {Didimo, Walter and Liotta, Giuseppe and Romeo, Salvatore},
year = {2010},
month = {09},
pages = {165-176},
title = {Topology-Driven Force-Directed Algorithms},
volume = {6502},
journal = {Proc. of GD 2010},
doi = {10.1007/978-3-642-18469-7_15}
}

@ARTICLE {zheng-gd2,
author = {J. X. Zheng and S. Pawar and D. M. Goodman},
journal = {IEEE Transactions on Visualization and Computer Graphics},
title = {Graph Drawing by Stochastic Gradient Descent},
year = {2019},
volume = {25},
number = {09},
issn = {1941-0506},
pages = {2738-2748},
doi = {10.1109/TVCG.2018.2859997},
publisher = {IEEE Computer Society},
address = {Los Alamitos, CA, USA},
month = {sep}
}

@ARTICLE{wang_2023_smartgd,
  author={Wang, Xiaoqi and Yen, Kevin and Hu, Yifan and Shen, Han-Wei},
  journal={IEEE Transactions on Visualization and Computer Graphics}, 
  title={Smart{GD}: A {GAN}-Based Graph Drawing Framework for Diverse Aesthetic Goals}, 
  year={2024},
  volume={30},
  number={8},
  pages={5666-5678},
  doi={10.1109/TVCG.2023.3306356}}

@misc{gephi,
  author       = {Gephi Consortium},
  title        = {Gephi},
  version      = {0.10.1},
  yearOPT         = {2023},
  howpublishedOPT = {Gephi Consortium},
  note         = {http://gephi.org}
}

@misc{graphviz,
  author       = {Graphviz},
  titleOPT        = {Graphviz},
  version      = {12.1.0},
  yearOPT         = {2024},
  howpublishedOPT = {Graphviz},
  note         = {http://graphviz.org}
}

@inproceedings{Mooney_2024_stress,
  title={The Perception of Stress in Graph Drawings},
  author={Gavin J. Mooney AND Helen Purchase AND Michael Wybrow  AND Stephen G. Kobourov AND Jacob Miller},
  booktitle={International Symposium Graph Drawing and Network Visualization},
  year={2024}
}

@inproceedings{Chimani_2014_stress,
  title={People Prefer Less Stress and Fewer Crossings},
  author={Markus Chimani AND Patrick Eades AND Peter Eades AND Seok-Hee Hong AND Weidong Huang AND Karsten Klein AND Michael Marner AND Ross T Smith AND and Bruce H Thomas},
  booktitle={International Symposium Graph Drawing and Network Visualization},
  year={2014}
}

@article{ye2024generative,
  author    = "{Ye}, {Yilin} and {Hao}, {Jianing} and {Hou}, {Yihan} and {Wang}, {Zhan} and {Xiao}, {Shishi} and {Luo}, {Yuyu} and {Zeng}, {Wei}",
  title    = "{Generative AI for Visualization: State of the Art and Future Directions}",
  journal   = "Visual Informatics",
  volume    = "8",
  issue     = "2",
  pages     = "43-66",
  year      = "2024",
  month     = "June"
}

@article{ScDiEl+2023Doom-1,
  author    = {Victor Schetinger and
               Sara Di Bartolomeo and
               Mennatallah El-Assady and
               Andrew Michael McNutt and
               Matthias Miller and
               Jane Lydia Adams},
  title     = {Doom or Deliciousness: Challenges and Opportunities for Visualization in the Age of Generative Models},
  journal   = {Eurographics Conference on Visualization (EuroVis)},
  volume    = {42},
  year      = {2023},
  doi       = {10.1111/cgf.14841}
}

@inproceedings{liu2021advisor,
  author    = {C. Liu and Y. Han and R. Jiang and X. Yuan},
  title     = {Advisor: Automatic visualization answer for natural-language question on tabular data},
  booktitle = {IEEE Pacific Visualization Symposium (PacificVis)},
  series    = {PacificVis '21},
  publisher = {IEEE},
  year      = {2021},
  pages     = {11--20},
  doi       = {},
  url       = {}
}

@article{dibia2019data2vis,
  author    = {V. Dibia and {\c{C}}. Demiralp},
  title     = {Data2vis: Automatic generation of data visualizations using sequence-to-sequence recurrent neural networks},
  journal   = {IEEE Computer Graphics and Applications},
  volume    = {39},
  number    = {5},
  pages     = {33--46},
  year      = {2019}
}

@article{wang2025aligned,
  title     = {How Aligned are Human Chart Takeaways and LLM Predictions? A Case Study on Bar Charts with Varying Layouts},
  author    = {Wang, Huichen Will and Hoffswell, Jane and Thane, Sao Myat Thazin and Bursztyn, Victor S. and Bearfield, Cindy Xiong},
  journal   = {IEEE Transactions on Visualization and Computer Graphics},
  volume    = {31},
  number    = {1},
  pages     = {536--546},
  year      = {2025},
  publisher = {IEEE},
  doi       = {10.1109/TVCG.2024.3456378},
  url       = {https://doi.org/10.1109/TVCG.2024.3456378}
}

@inproceedings{DiBartolomeo2023AskYouShall,
  author    = {Di Bartolomeo, Sara and Severi, Giorgio and Schetinger, Victor and Dunne, Cody},
  title     = {Ask and You Shall Receive (a Graph Drawing): Testing ChatGPT's Potential to Apply Graph Layout Algorithms},
  booktitle = {EuroVis 2023 - Short Papers},
  year      = {2023},
  editor    = {H{\"o}llt, Thomas and Aigner, Wolfgang and Wang, Bei},
  publisher = {The Eurographics Association},
  isbn      = {978-3-03868-219-6},
  doi       = {10.2312/evs.20231047},
  url       = {https://osf.io/n5rxd/}
}

@inproceedings{grover2016node2vec,
  title     = {node2vec: Scalable Feature Learning for Networks},
  author    = {Aditya Grover and Jure Leskovec},
  booktitle = {Proceedings of the 22nd ACM SIGKDD International Conference on Knowledge Discovery and Data Mining (KDD)},
  year      = {2016},
  pages     = {855--864},
  publisher = {ACM},
  doi       = {10.1145/2939672.2939754}
}

@article{Eades2017ShapeMetrics,
  author    = {Peter Eades and Seok-Hee Hong and An Nguyen and Karsten Klein},
  title     = {Shape-Based Quality Metrics for Large Graph Visualization},
  journal   = {Journal of Graph Algorithms and Applications},
  volume    = {21},
  number    = {1},
  pages     = {29--53},
  year      = {2017},
  doi       = {10.7155/jgaa.00405},
  url       = {http://jgaa.info/getPaper?id=405},
}

@inproceedings{Cai2021PacificVis,
  author    = {Shijun Cai and Seok‐Hee Hong and Jialiang Shen and Tongliang Liu},
  title     = {A Machine Learning Approach for Predicting Human Preference for Graph Layouts},
  booktitle = {2021 IEEE 14th Pacific Visualization Symposium (PacificVis)},
  year      = {2021},
  pages     = {6--10},
  organization = {IEEE}
}

@incollection{Klammler2018GD,
  author    = {Moritz Klammler and Tamara Mchedlidze and Alexey Pak},
  title     = {Aesthetic Discrimination of Graph Layouts},
  booktitle = {Graph Drawing and Network Visualization: 26th International Symposium, GD 2018, Barcelona, Spain, September 26–28, 2018, Proceedings},
  publisher = {Springer International Publishing},
  year      = {2018},
  pages     = {169--184},
  address   = {Cham, Switzerland}
}

@article{Wu2025GNNGraph,
  author    = {Xiangyang Wu and Qian Li and Xiaodong Pan and Xiaozhi Liu and Zhen Liu},
  title     = {A Deep Learning Approach to Evaluate the Quality of Graph Layouts Using {GNN}},
  journal   = {Journal of Visualization},
  volume    = {28},
  number    = {2025},
  pages     = {413--429},
  year      = {2025}
}

@inproceedings{Wu2021CHIChart,
  author    = {Aoyu Wu and Liwenhan Xie and Bongshin Lee and Yun Wang and Weiwei Cui and Huamin Qu},
  title     = {Learning to Automate Chart Layout Configurations Using Crowdsourced Paired Comparison},
  booktitle = {Proceedings of the 2021 CHI Conference on Human Factors in Computing Systems},
  year      = {2021},
  pages     = {1--13}
}

@misc{siméoni2025dinov3,
      title={DINOv3}, 
      author={Oriane Siméoni and Huy V. Vo and Maximilian Seitzer and Federico Baldassarre and Maxime Oquab and Cijo Jose and Vasil Khalidov and Marc Szafraniec and Seungeun Yi and Michaël Ramamonjisoa and Francisco Massa and Daniel Haziza and Luca Wehrstedt and Jianyuan Wang and Timothée Darcet and Théo Moutakanni and Leonel Sentana and Claire Roberts and Andrea Vedaldi and Jamie Tolan and John Brandt and Camille Couprie and Julien Mairal and Hervé Jégou and Patrick Labatut and Piotr Bojanowski},
      year={2025},
      eprint={2508.10104},
      archivePrefix={arXiv},
      primaryClass={cs.CV},
      url={https://arxiv.org/abs/2508.10104}, 
}

@article{gower1975generalized,
  title={Generalized procrustes analysis},
  author={Gower, John C},
  journal={Psychometrika},
  volume={40},
  number={1},
  pages={33--51},
  year={1975},
  publisher={Springer-Verlag}
}

\clearpage
\appendix

\section{Additional Details on User Preference Label Collection}

We constructed a large-scale labeled dataset through a carefully designed user study. The dataset comprises visualizations of 11{,}531 unique graphs, each rendered using eight canonical layout algorithms. Human participants were asked to compare the layout variants for each graph and select the one they perceived as most aesthetically pleasing.

\subsection{Candidate Graphs and Layout Algorithms}
The graph dataset used in this study is the \textit{Rome Graphs Collection} (\url{http://www.graphdrawing.org/data.html}), a standard benchmark dataset in the graph drawing literature. It comprises 11,531 small to medium-sized graphs drawn from a variety of application domains. Each graph consists of on average 52 vertices (min = 10, max = 110) and 69 edges (min = 9, max = 158), making them suitable for human visual inspection and preference judgment.

To investigate aesthetic preferences across a diverse range of layout techniques, we rendered each graph using eight commonly used layout algorithms, including 
\texttt{Neato}~\cite{neato} (based on stress minimization from GraphViz~\cite{graphviz});  
\texttt{Kamada-Kawai (KK)}~\cite{kamada_kawai_1989} (classical spring model); 
\texttt{ForceAtlas2 (FA2)}~\cite{fa2} (force-directed layout model from Gephi~\cite{gephi}).
\texttt{fdp} (force-directed layout algorithm  from GraphViz~\cite{graphviz});
\texttt{sfdp}~\cite{sfdp} (multilevel force-directed layouts  from GraphViz);
\texttt{spring}~\cite{spring} (NetworkX implementation of Fruchterman-Reingold force-directed algorithm);
\texttt{PMDS}~\cite{pmds} (fast multidimensional scaling); and
\texttt{spectral}~\cite{spectral} (NetworkX implementation of graph layout using eigenvectors of the graph Laplacian).

These eight algorithms span several layout paradigms, including force-directed, spectral, and multidimensional scaling, ensuring visual diversity across layout candidates.

\subsection{User Study Design Principles and User Interface}

Our data collection was guided by three core principles: (1) \textit{Scalability}, to support annotation of thousands of graphs; (2) \textit{Coverage and Redundancy}, to ensure label robustness through repeated labeling; and (3) \textit{Conflict Resolution}, to handle disagreement in subjective preferences.



\begin{figure}[htbp!]
        \includegraphics[width=0.5\textwidth]{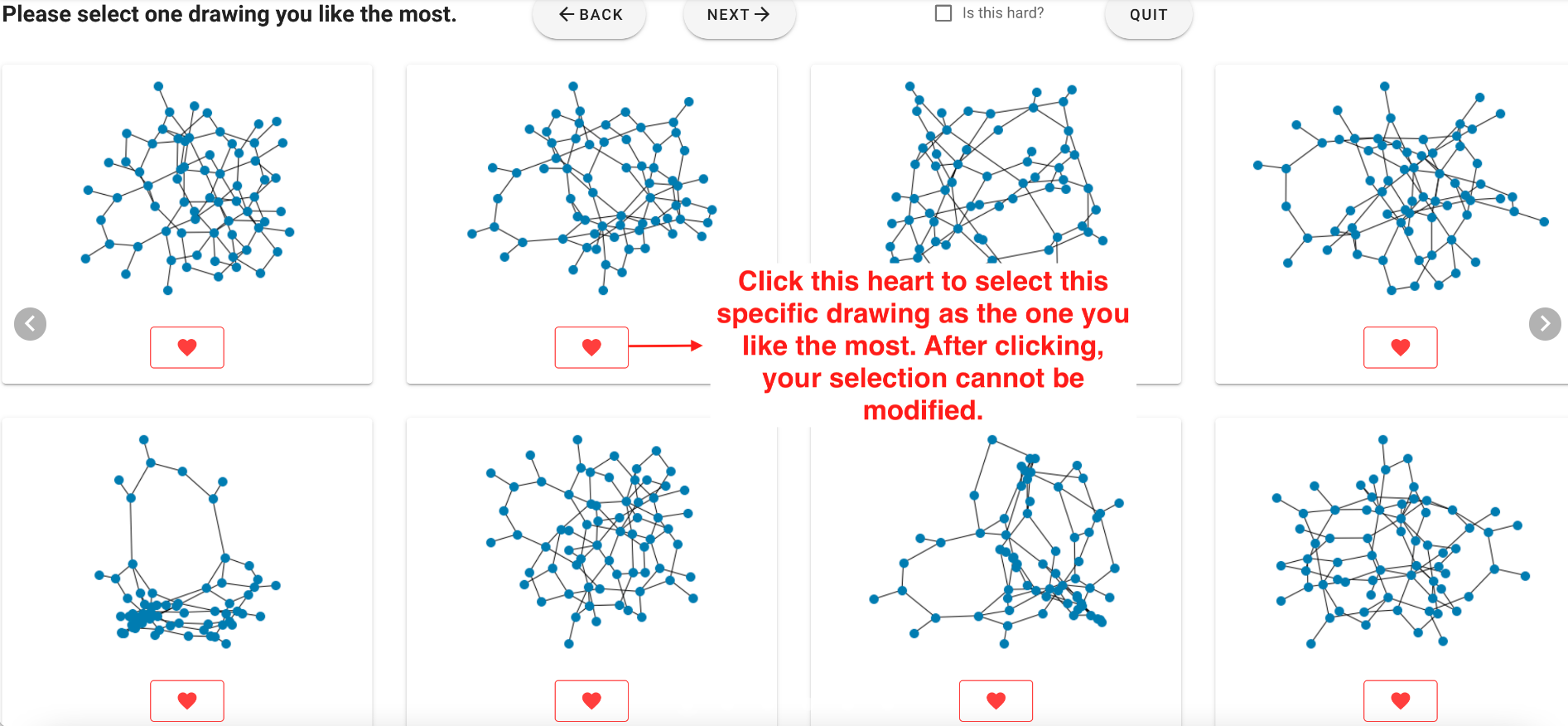} 
    \caption{A screenshot showing the training provided to the participants}
    \label{fig:training}
\end{figure}

Participants labeled graphs through a web interface~(Figure~\ref{fig:training}) in a within-subject study. In each task, they were shown eight visualizations of the same graph, each produced by a different canonical layout algorithm. The eight layouts were displayed simultaneously in a randomly ordered grid, without any algorithm labels to avoid bias. Participants selected the layout they found most visually pleasing and could optionally note if the choice was difficult. We also recorded the time spent on each task as a proxy for decision effort. The interface emphasized accessibility and low cognitive load, allowing both casual and experienced users to participate easily.

\subsection{Participant Training}

After creating an account, each participant is guided through a brief training tutorial session. First, a message is shown to explain the purpose of the study: \textit{``We are conducting a research project related to graph drawing.  
We hope you can help us build a collection of human-preferred graph layouts. For each graph, you will be asked to choose the one drawing you like the most from a set of 8. There is no required number of graphs to label—we simply ask that you label as many as you can.  
Please take a moment to review the user manual below before you begin.''}

Participants are then shown a sequence of annotated images introducing the interface. These annotations highlight clickable buttons. For example, one instruction reads: ``Click this heart to select the drawing you like the most.'' This tutorial was made available at any time via a prominently placed button ``tutorial''.

\subsection{Labeling Tasks}

The participant's task was as follows: 1) visually inspect all eight layout variants of the same graph and elect the layout they found \emph{most aesthetically pleasing} by clicking a heart-shaped button; 2) optionally, check a box labeled \textit{``Is this hard?''} if the decision was perceived as difficult or ambiguous; 
Time spent on the task was automatically recorded from the moment the layout grid was shown until a selection was made.

To keep participants informed and engaged, the interface provided visual progress feedback with a running total of the number of graphs the participant has labeled. Even 50 labels, the UI will display a motivational popup message such as \textit{``Good job! You have labeled 8,350 graphs.  
You have labeled more graphs than 85.71\% of users. Please keep up the great work!''}

\subsection{Adaptive Graph Assignment Logic}

Each participant was sequentially assigned graphs according to a priority system designed to ensure broad coverage and reliable labeling. Graphs with no prior labels were presented first to maximize coverage across the dataset. Next, graphs with the fewest label were prioritized to strengthen agreement estimates.


To reduce annotation fatigue and maintain engagement, participants could skip difficult tasks. Skipped graphs were placed in a personal queue and periodically resurfaced during the session with a fixed probability (40\%), encouraging eventual completion while respecting user agency. Participants could also revisit previously skipped items in order, supporting flexibility and reflection. Further implementation details are provided in the Appendix.




\subsection{Labeled Data}

We recruited 27 participants. 23 of the participants are complete novices; 2 of the participated has some knowledge of graph visualization, and 2 of them are expert in the area. The labeling was done in a period spanning May 2022 to August 2025.
In total we collected 64,436 labels; on average, each graph is labelled 5.58 times. On average, the participants took 9.61 seconds per label.

\begin{figure}[htbp!]
        \includegraphics[width=0.5\textwidth]{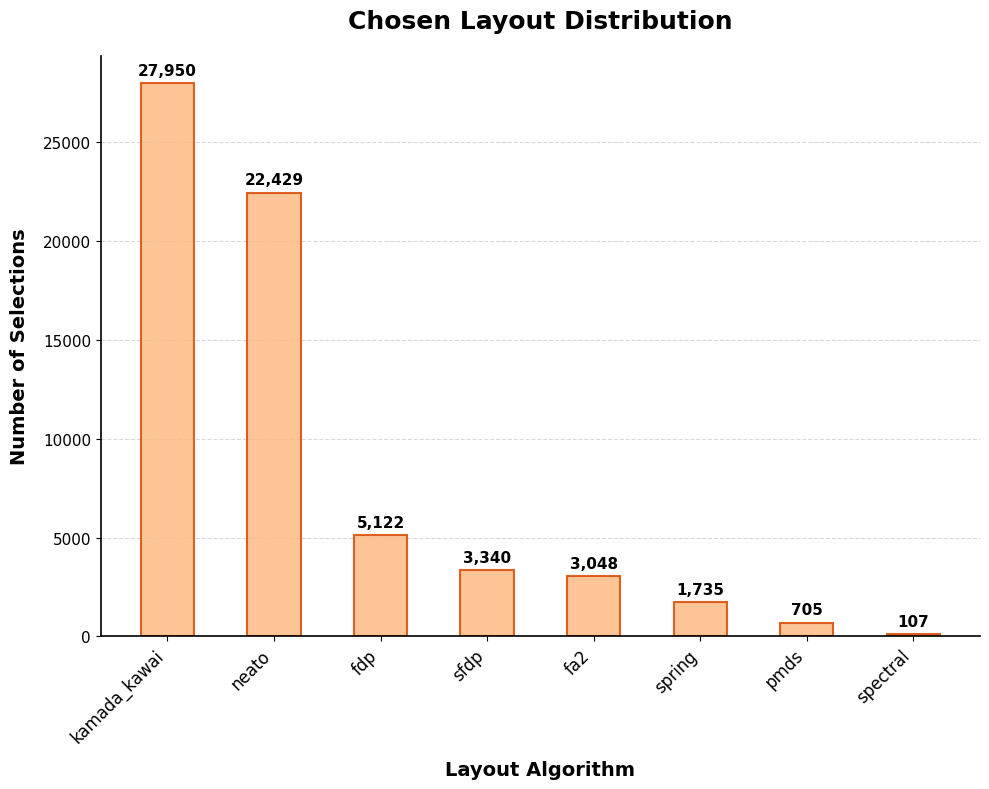} 
    \caption{Distribution of layout algorithms behind the visualizations chosen by human subjects as being the most pleasing}
    \label{fig:layout_distribution}
\end{figure}

We found that the visualizations from different algorithms have different probabilities of being selected. Figure~\ref{fig:layout_distribution} show the counts of each chosen layout. Among the eight layouts, kamada\_kawai is the most frequently selected (44.3\%), followed by neato (34.8\%). The remaining layouts each account for less than 10\% of the labels, with spectral being rarely chosen (0.17\%). Such an imbalance is important as it reflects human preferences.

Each graph received between three and seven labels, enabling analysis of human–human alignment. Agreement was defined as the proportion of commonly labeled graphs on which two users selected the same layout. 
Overall, agreement rates were moderate: most user pairs fell within the 28\%–51\% range, indicating partial alignment but substantial diversity in preferences. The highest agreement occurred between Users~104 and~102 (50\%), while the lowest was between Users~1 and~108 (29

This modest alignment is unsurprising, as several layout algorithms often produce visually similar results. For instance, Neato and Kamada–Kawai layouts tend to generate comparable structures, which likely explains why both received the most user ``votes.''

\subsection{Human feedback on their selection criteria}

We conducted an informal survey of study participants, asking them to explain why they preferred one layout over another. We then used a large language model (LLM) to analyze their free-form feedback and summarize the most common reasons for liking or disliking a visualization, as shown in Tables~\ref{tab:like_reasons}-\ref{tab:dislike_reasons}. As can be seen, human prefers symmetry/geometry harmony/clear, balanced, and clean structure, less edge crossing, and uniform edge lengths. They do not like crowded/cluttered areas, node and edge overlaps, and distorted shapes.

\begin{table}[htbp]
\centering
\caption{Top reasons for liking a visualization \& frequency}
\label{tab:like_reasons}
\setlength{\tabcolsep}{1pt} 
\begin{tabular}{>{\raggedright\arraybackslash}p{2.5cm} >{\raggedright\arraybackslash}p{5cm} c}
\toprule
\textbf{Reason} & \textbf{Description} & \textbf{Freq} \\
\midrule
Symmetry & Layouts with symmetric arrangements, balanced proportions, or bilateral harmony. & 42 \\
No edge crossings & Absence of edge intersections or overlaps, leading to planar and clean visuals. & 35 \\
Clear structure & Readable, clear overall structure or preserved graph shape. & 31 \\
Good spacing & Even node spacing, well-spread layout, or efficient use of space. & 29 \\
Geometric harmony & Harmonious proportions, stability, or appealing geometric metaphors (e.g., cubic, spherical). & 25 \\
Uniform edge lengths & Consistent or even edge lengths without variation. & 22 \\
Visually clean & Clean, uncluttered appearance with minimal visual noise. & 20 \\
Smooth and natural shape & Smooth curves, natural flow, or organic/appealing forms. & 18 \\
Preserved central structure & Well-maintained core or hub elements without distortion. & 17 \\
Balanced angular resolution & Good angles between edges, avoiding tight or distorted angles. & 15 \\
\bottomrule
\end{tabular}
\end{table}

\begin{table}[htbp]
\centering
\caption{Top reasons for disliking a visualization \& frequency}
\label{tab:dislike_reasons}
\setlength{\tabcolsep}{0pt} 
\begin{tabular}{>{\raggedright\arraybackslash}p{2.5cm} >{\raggedright\arraybackslash}p{5cm} c}
\toprule
\textbf{Reason} & \textbf{Description} & \textbf{Freq} \\
\midrule
Asymmetry or imbalance & Lopsided, unbalanced, or asymmetrical layout. & 45 \\
Edge crossings & Presence of intersecting or crossing edges causing ambiguity. & 38 \\
Dense or crowded areas & Clustered, dense, or congested regions with too much proximity. & 36 \\
Overlapping edges/nodes & Edges or nodes overlapping, leading to confusion or invisibility. & 32 \\
Distorted or skewed shape & Warped, distorted, or irregular overall shape. & 26 \\
Uneven edge lengths & Inconsistent, irregular, or varying edge lengths. & 24 \\
Poor spacing & Uneven, poor, or inadequate spacing between elements. & 23 \\
Stretched or elongated & Overly stretched, elongated, or disproportionate sections. & 21 \\
Cluttered or messy & Chaotic, messy, or cluttered appearance with visual noise. & 19 \\
Sharp or tight angles & Sharp turns, tight angles, or poor angular resolution. & 16 \\
\bottomrule
\end{tabular}
\end{table}

\section{Additional Results Analysis}

\subsection{Similarity-aware Alignment}

\revision{
One observation from the dataset is that certain layout algorithms can produce visually similar drawings for the same graph. 
Such similarities may introduce ambiguity for participants during labeling and, consequently, may influence the alignment calculation. To address this issue, we introduce a similarity-aware alignment metric that incorporates a similarity threshold as a parameter. 
Let the similarity between two layouts be defined using Procrustes analysis \cite{gower1975generalized}. 
Given two layouts $X, Y \in \mathbb{R}^{n \times 2}$ of the same graph, we compute their Procrustes distance $ d_{\text{proc}}(X, Y) $ and we define the similarity score as
\begin{equation}
S(X, Y) = 1 - d_{\text{proc}}(X, Y),
\end{equation}
after appropriate normalization of $d_{\text{proc}}$ to $[0,1]$. 
Given a predefined threshold $\alpha$, two layouts are considered equivalent if

\begin{equation}
S(X, Y) \ge \alpha.
\end{equation}

Under this definition, the similarity-aware pairwise alignment between labelers $i$ and $j$ is defined as

\begin{equation}
\text{Alignment}_{\alpha}(i, j)
=
\frac{
\sum_{G \in D(i) \cap D(j)}
\delta \big(
S(l(G,i),\, l(G,j)) \ge \alpha
\big)
}{
|D(i) \cap D(j)|
},
\end{equation}

\noindent and the corresponding micro-averaged alignment is defined as

\begin{equation}
\text{Alignment}_{\alpha}
=
\frac{
\sum_{i,j}
\sum_{G \in D(i) \cap D(j)}
\delta \big(
S(l(G,i),\, l(G,j)) \ge \alpha
\big)
}{
\sum_{i,j}
|D(i) \cap D(j)|
}.
\end{equation}
}

\subsection{Statistic Analysis based on Similarity-aware Alignment}

\revision{
We further evaluate model performance under the similarity-aware 
alignment metric defined in Equations (1)–(4). Table~\ref{tab:agreement_threshold} reports 
micro-averaged agreement across varying similarity thresholds 
$\alpha$. Notably, at 
$\alpha = 0.95$, VM--User agreement reaches 50.67\%, which is 
comparable to Human--Human agreement (50.11\%), suggesting that 
the vision model approaches the level of inter-human consistency 
under high-similarity constraints. These results indicate that accounting for layout similarity provides 
a more nuanced understanding of alignment. When visually indistinguishable layouts are treated as equivalent, agreement levels among human–human, human–VM, and human–LLM comparisons increase substantially. This suggests that human visual preferences are far from random. Although individual preferences vary, Table~\ref{tab:agreement_threshold} also indicates the presence of a degree of universality.}

\begin{table}[ht]
\centering
\caption{\revision{$Agreement_{\alpha}$ Summary at Different Thresholds}}
\label{tab:agreement_threshold}
\begin{tabular}{cccc}
\hline
\textbf{Threshold} & 
\textbf{User--User (\%)} & 
\textbf{VM--User (\%)} & 
\textbf{LLM--User (\%)} \\
\hline
0.00 & 100.00 & 100.00 & 100.00 \\
0.50 & 96.17  & 96.87  & 95.48  \\
0.60 & 90.71  & 92.51  & 88.36  \\
0.70 & 83.25  & 84.69  & 79.67  \\
0.80 & 71.66  & 72.50  & 67.64  \\
0.85 & 65.83  & 67.09  & 61.47  \\
0.90 & 58.99  & 60.18  & 53.86  \\
0.95 & 50.11  & 50.67  & 44.63  \\
0.96 & 48.45  & 49.16  & 42.74  \\
0.97 & 46.95  & 47.08  & 41.13  \\
0.98 & 45.21  & 44.66  & 39.64  \\
0.99 & 43.05  & 42.15  & 37.52  \\
1.00 & 38.66  & 36.82  & 33.11  \\
\hline
\end{tabular}
\end{table}

\subsection{Examples of Large Graph}

\begin{figure}[htbp!]
        \includegraphics[width=0.5\textwidth]{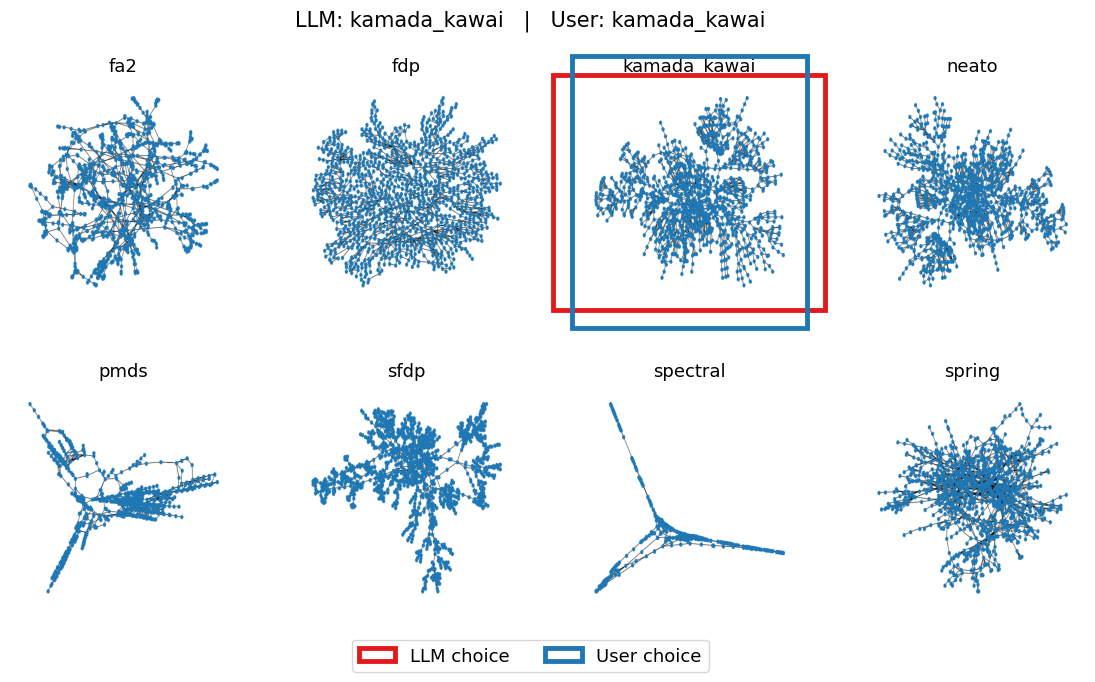} 
    \caption{\revision{A screenshot showing the preference of humans vs that of the LLM for graph 1138\_bus ($|V|=1138, |E|=1452$).}}
    \label{fig:bus_llm_user}
\end{figure}

\begin{figure}[htbp!]
        \includegraphics[width=0.5\textwidth]{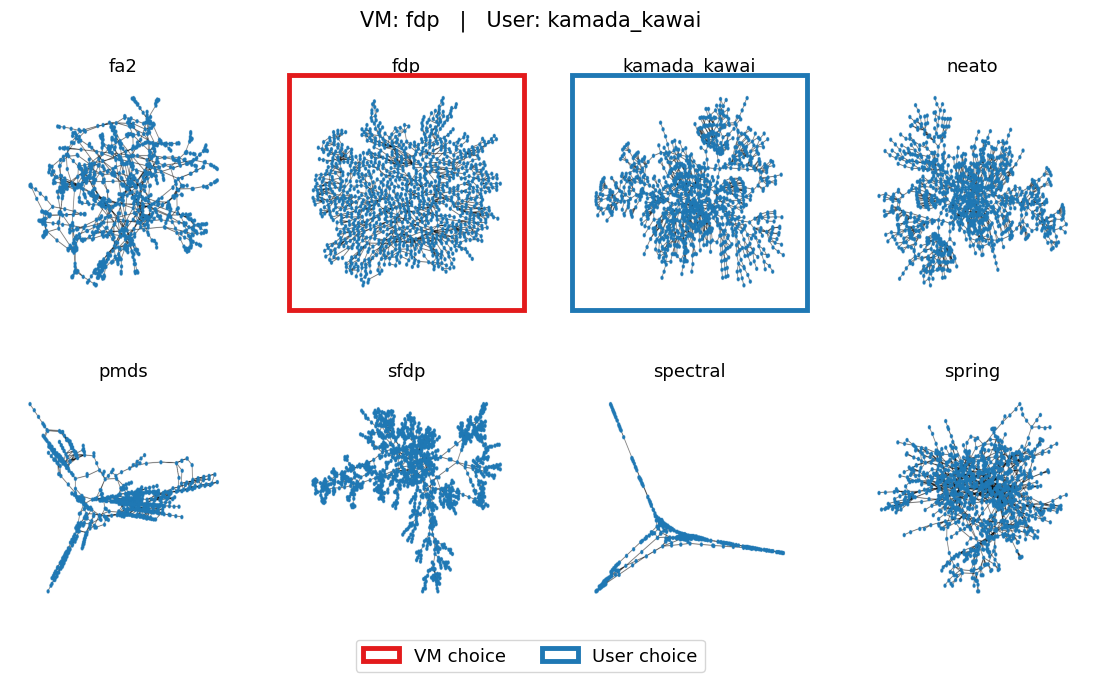} 
    \caption{\revision{A screenshot showing the preference of human vs that of the VM for graph 1138\_bus($|V|=1138,|E|=1452$)}}
    \label{fig:bus_vm_user}
\end{figure}

\begin{figure}[htbp!]
        \includegraphics[width=0.5\textwidth]{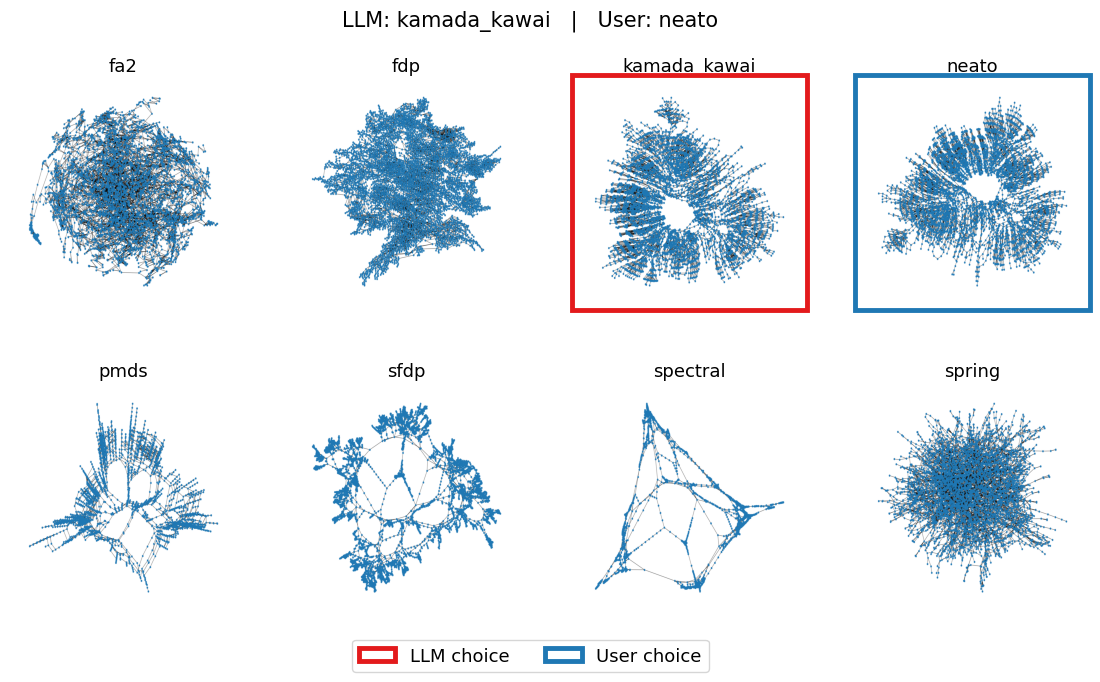} 
    \caption{\revision{A screenshot showing the preference of human vs that of the LLM for graph USPowerGrid($|V|=4591,|E|=6594$)}}
    \label{fig:power_llm_user}
\end{figure}

\begin{figure}[htbp!]
        \includegraphics[width=0.5\textwidth]{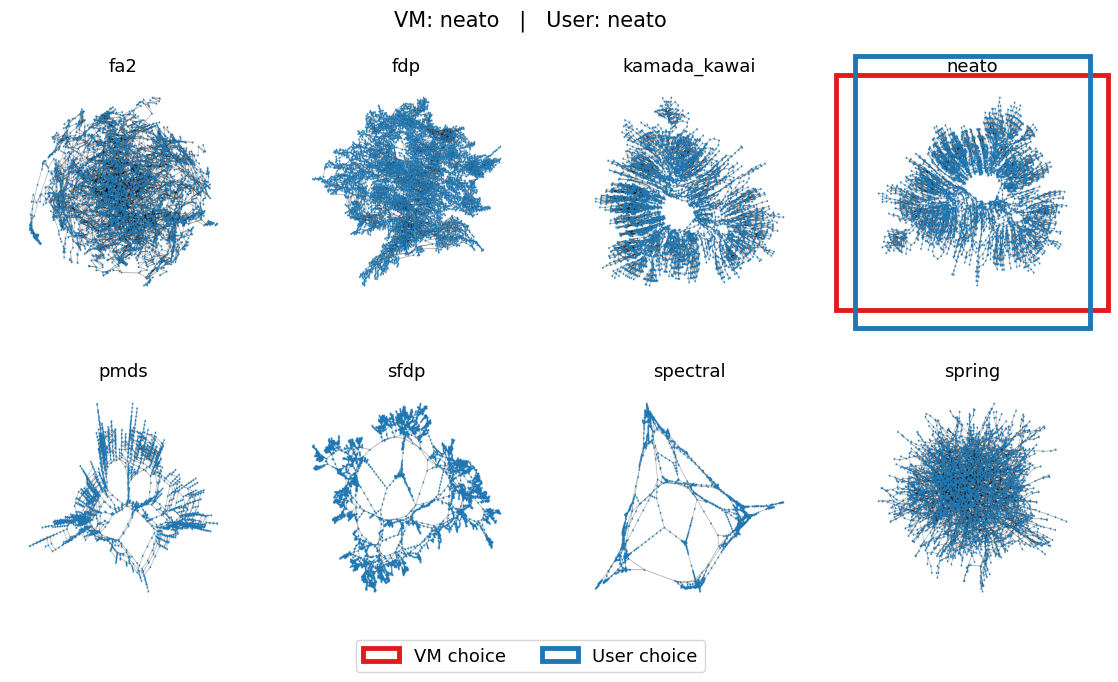} 
    \caption{\revision{A screenshot showing the preference of human vs that of the VM for graph USPowerGrid ($|V|=4591,|E|=6594$)}}
    \label{fig:power_vm_user}
\end{figure}

\revision{
Figures \ref{fig:bus_llm_user} - \ref{fig:power_vm_user} present representative examples of layout comparisons 
for larger graphs. In each example, we highlight the layout selected 
by the human annotator and the layout chosen by the model. These examples illustrate two typical scenarios: (1) cases where the 
model agrees with the human selection, often choosing visually 
balanced and symmetric layouts; and (2) disagreement cases. Overall, qualitative inspection suggests that the vision model 
captures several aesthetic principles reported by participants, such 
as symmetry, balanced spacing, and reduced edge crossings.
}

\subsection{Statistical Significance Tests}

\revision{
To evaluate whether observed performance differences are statistically 
meaningful, we conducted paired-sample t-tests using per-user 
alignment scores.

For the comparison across different model sizes, the paired-sample 
t-test indicated that the difference was not statistically significant 
($t = 1.1245$, $p = 0.2608$). This suggests that increasing model 
size does not lead to a statistically reliable improvement in alignment 
performance in our setting.

For the backbone comparison reported in Table~8, DINOv2 achieved 
significantly higher alignment scores than ResNet-50. A paired-sample 
t-test confirmed that this improvement is statistically significant 
($t = 2.1483$, $p = 0.0317$, $p < 0.05$). This result indicates that 
DINOv2 more effectively approximates human layout preferences 
compared to ResNet-50.
}

\subsection{Model Performance on All-human-agreed subsets}

\revision{ we evaluate performance on the 
subset of graphs for which all human annotators agreed on the same 
layout. This subset represents instances with strong and unambiguous 
aesthetic preference. This subset contains 45 graphs. On this subset, the VM achieves an 
alignment of 28.89\%, while the LLM achieves 40.00\%. Although the LLM shows higher alignment on this subset, the small 
sample size limits the statistical power of this comparison.
}

\end{document}